\documentclass{article}

\usepackage{arxiv}

\usepackage[utf8]{inputenc} 
\usepackage[T1]{fontenc}    
\usepackage{hyperref}       
\usepackage{url}            
\usepackage{booktabs}       
\usepackage{amsfonts}       
\usepackage{nicefrac}       
\usepackage{microtype}      
\usepackage{lipsum}
\usepackage{comment}
\usepackage{epsfig}
\usepackage{graphicx}
\usepackage{amsmath}
\usepackage{amssymb}
\usepackage{multicol}
\usepackage{booktabs}
\usepackage{tabu}
\usepackage{multirow}
\usepackage{wrapfig}
\usepackage{verbatim}
\usepackage{caption}
\usepackage{subcaption}
\usepackage{setspace}

\title{AnoNet: Weakly Supervised Anomaly Detection in Textured Surfaces}

\author{
  Manpreet Singh Minhas, John Zelek \\
  Department of Systems Design Engineering\\
  University of Waterloo\\
  Waterloo, ON, Canada\\
  \texttt{\{msminhas,jzelek\}@uwaterloo.ca} \\
}

\begin{document}
\maketitle


\begin{abstract}

Humans can easily detect a defect (or anomaly) whether it be a scratch on a car door or a crack on the road.
This anomaly stands out because it is different or salient when compared to the surface it resides on.
Today, manual human visual inspection is still the norm because it is difficult to automate anomaly detection and classification. Deep learning neural networks are a useful tool that can teach a machine to find defects. The problem is that deep networks require a lot of training examples to learn what a defect is and it is tedious and expensive to get these samples. We tackle the problem of teaching a network with a low number of training samples with a system we call AnoNet.  AnoNet's architecture is similar to CompactCNN with the exceptions that (1) it is a fully convolutional network and does not use strided convolution; (2) it is shallow and compact which minimizes over-fitting by design; (3) the compact design constrains the size of intermediate features which allows training to be done without image downsizing; (4) the model footprint is low making it suitable for edge computation; and (5) the anomaly can be detected and localized despite the weak labelling. AnoNet learns to detect the underlying shape of the anomalies despite the weak annotation as well as preserves the spatial localization of the anomaly. 
Pre-seeding AnoNet with an engineered filter bank initialization technique reduces the number of samples required for training and also achieves state-of-the-art-performance.
Compared to the CompactCNN, the AnoNet architecture achieved a massive 94\% reduction of network parameters from 1.13 million to 64 thousand parameters.
Experiments were conducted on four data-sets and results were compared against  CompactCNN and DeepLabv3. AnoNet improved the performance on an average across all data-sets by 106\% to an F1 score of 0.98 and by 13\% to an AUROC value of 0.942. AnoNet can learn from a limited number of images. For one of the data sets, AnoNet learnt to detect anomalies after a single pass through just 53 training images. 
\end{abstract}

\section{Introduction}
According to the World Health Organization (WHO) Global Status Report on Road Safety 2018, there are 1.35 million road traffic deaths every year \cite{whoreport}. A study conducted by the Pacific Institute for Research and Evaluation (PIRE) on traffic accidents and fatalities in 2009, found that more than half of the deaths that occurred on the American roadways were due to poor road conditions \cite{pirereport}. The expense of those accidents costs the U.S. economy $\$217$ billion each year. Additionally, $\$91$ billion was invested annually in road infrastructure  \cite{roadexpense}. Poor road conditions are primarily caused due to surface defects such as cracks and potholes. In the railways, broken rails and welds were the most common cause of train derailments which accounted for more than $15\%$ of defects \cite{railaccidents}. Detecting defects in industrial manufacturing processes is crucial for ensuring the high quality of finished products. All these examples show the vital importance of detecting defects across different industries. 

A common property of these surface defects is that their visual texture is inherently different from the defect-free surface. Since human visual inspection relies solely on what is seen, it only makes sense that automating visual inspection from camera images should be plausible. The task of automated visual defect detection can, therefore, be formulated as the problem of anomaly detection in textured surfaces. Visual texture refers to the human visual cognition and semantic meaning of textures based on the local spatial distributions of simple appearance properties such as color, structure, reflectance, and orientation of the object. The objective of the manual human inspection is to detect anomalies by comparing the difference in the visual texture appearance of the defects against defect free appearance. The process is not only time consuming and expensive but also prone to errors due to the monotony of the task. It is also subjective and susceptible to human biases. Individual factors such as age, visual acuity, gender, experience, scanning strategy, training, etc. also affect inspection \cite{see_inspection}. To overcome these problems and challenges, a significant amount of work has been conducted to automate the process of anomaly detection in textured surfaces. Examples of automatic visual inspection systems in various domains include defect detection in steel surfaces \cite{app8112195}, pavements \cite{pavement}, rail tracks \cite{railtrack} and fabric \cite{fabricdefect} to name a few. The key challenges faced by automated detection systems are as follows. Anomalies such as dents, smudges, cracks, impurities, scratches, stains, etc. vary in terms of pixel intensities, geometric constraints, and visual appearance \cite{compactcnn}. Moreover, the data often contains noise, which although is different from the normal data, should not be classified as an anomaly by the detection system. Environmental factors such as lighting, temperature, extreme weather (such as snow) also impact the detection systems. These challenges make anomaly detection in textured surfaces extremely complex and difficult.

For these automated systems, textures can be described using two approaches, namely structural and statistical. The structural approach considers texture as an organised area phenomenon which can be decomposed into primitives (also known as textons) having specific spatial distributions \cite{haralick}. This definition comes directly from the human visual experience of textures. The number and type of primitives, as well as their spatial organization or layout, describes an image texture. For example, a brick wall texture is generated by tiling up bricks (primitives or textons) in a layered manner (specific spatial distribution). The second approach known as the statistical approach considers textures to be generated by a stochastic process such as a Markov Random Field. Grass, sand, sandpaper, leather, etc. are examples of this category. The quantitative measure of the spatial distribution of gray levels in textured surfaces forms the basis of the statistical approach.

Traditionally, the automated methods have relied on the computation of a set of hand-crafted textural features in the spatial or spectral domain followed by the search for significant deviations in the feature values by using different classifiers. In spatial-domain approaches, the commonly used features are second-order statistics derived from spatial gray-level co-occurrence matrices \cite{TSAI2003307}. Spectral approaches normally involve the use of Gabor filters \cite{gabor}, Fourier transform \cite{fourierdefect} and Wavelet transform \cite{waveletdefect}. These methods, however, suffer from following drawbacks. The hand-crafted features require domain expertise and are very challenging to formulate. They do not generalize which means that the engineered features that are designed for a specific task cannot be used for other different or even similar tasks. 

Deep learning techniques applied to anomaly detection have overcome these challenges and are receiving increased attention. Convolutional Neural Networks (CNNs) were used for supervised steel defect classification \cite{Masci2012SteelDC} and rail surface defect classification \cite{roohi}. Although the deep learning techniques have outperformed the traditional hand-crafted features based approaches, they suffer from their own associated set of challenges. The lack of available labelled training examples is a major challenge \cite{chalapathyanomaly} \cite{chandola2009anomaly}, since these models require large labelled data-sets. The instances of anomalous classes are even fewer in these data-sets which hinders the training of the network. Pixel-level annotated data-sets that are required for supervised defect segmentation, are not only rarer but also expensive and time-consuming. 

Weakly supervised learning covers techniques that try to construct models by learning with weak supervision and directly addresses these pain points. Weak supervision can be broadly classified into three categories: incomplete, inexact and inaccurate \cite{zhouweaklysupervised}. In incomplete supervision only a small subset of training set has labels and the rest of the samples are unlabelled. Inexact supervision involves coarse-grained labels, for example, image-level labels rather than object-level labels. Inaccurate supervision involves labels that are not always correctly labelled. One technique that is in both the inexact and inaccurate category is weakly supervised anomaly detection. It uses masks that are loosely annotated at the pixel level e.g. masks in the form of some geometric shape such as an ellipse covering the entire anomaly. As a result, the mask only provides a coarse location of the anomaly and there are a lot of inaccurately labelled normal (defect-free) background pixels which are seen as anomalous pixels. This makes the anomaly detection task even more challenging.

In this paper, we present a novel technique for anomaly detection in textured surfaces using weakly annotated data, that is capable of learning the underlying shape of the anomaly from not only weakly annotated data but also from a limited number of examples.

\newpage

\begin{figure*}[!ht]
\begin{center}
   \includegraphics[width=\linewidth]{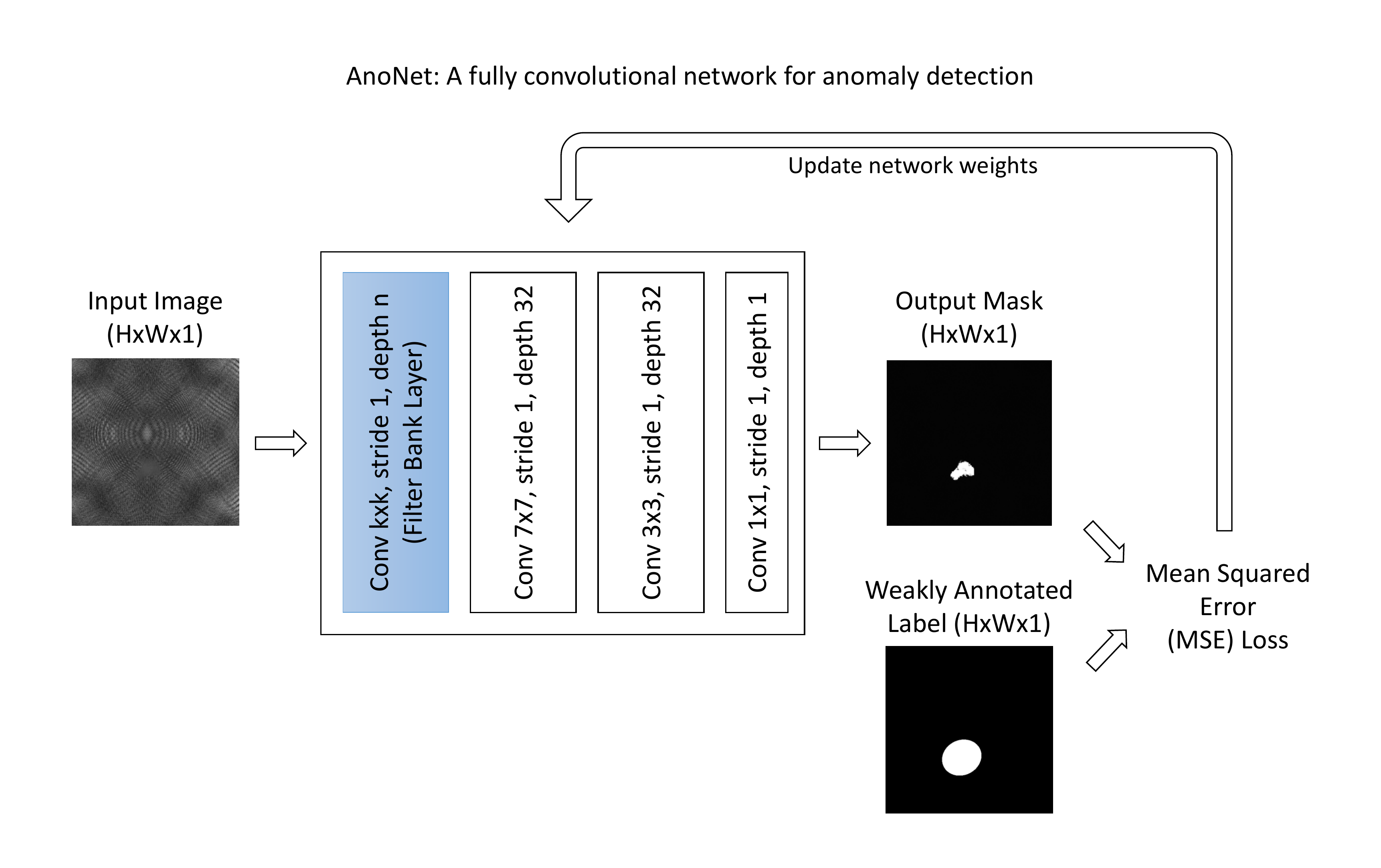}
\end{center}
   \caption{AnoNet: A fully convolutional network, for anomaly detection in textured surfaces using weakly labelled data that outputs a $H\times W$ segmentation mask for a $H\times W$ input image. With only 64 thousand parameters, AnoNet is remarkably compact and not susceptible to the problem of over-fitting. It has the valuable and important ability to learn to detect the actual shape of the anomaly not only from the weak annotations but also from a limited number of samples. It uses a filter bank initialization technique for the first layer, as described in \ref{sec:filterbankinitialization} and the values of network parameters $k$(filter size) and $n$ (filter stack length) depend on the filter bank being used and are described in subsection \ref{sec:filterbankinitialization}. For the rest of the network the weights are initialized using a random normal distribution of mean zero and variance of one.}
\label{fig:anonetoverview}
\end{figure*}

Our contributions are as follows:
\begin{enumerate}
    \item We have presented AnoNet (Figure \ref{fig:anonetoverview}), a fully convolutional architecture with only 64 thousand parameters, for anomaly detection in textured surfaces using weakly labelled data that outputs a $H\times W$ segmentation mask for a $H\times W$ input image. This prevents the loss of localization of the anomaly with respect to the original image. The network has a valuable and important ability to learn to detect the actual shape of the anomaly despite the weak annotations.
    \item AnoNet has an important practical advantage for real-world applications.  It can learn from a limited number of weakly annotated images. For the RSDDs-I data-set, it learnt to detect anomalies after just a single pass through a small sized training set of 53 images.
    \item A filter bank based initialization technique for AnoNet is presented. To the best of our knowledge, no such work has been done for weakly supervised anomaly detection in textured surfaces. AnoNet achieved state of the art performance on four challenging data-sets. In comparison to the CompactCNN \cite{compactcnn} and DeepLabv3 \cite{deeplabv3}, AnoNet achieved an impressive average improvement in performance by 106\% to an F1 score of 0.98 and by 13\% to an AUROC value of 0.942. We hypothesize that alleviating the network for learning features focuses learning on better discriminators of anomalies.
\end{enumerate}

\section{Related Work}
\label{sec:relatedwork}

The primary goal of defect detection and assessment is to differentiate possible defective regions from non-defective regions. For visual appearance inspection, this is just the task of anomaly detection in textured surfaces. Different and complex textures, diverse shapes, sizes and colors of defects, as well as varying lighting conditions, make the task extremely challenging. Traditional methods follow the pipeline of feature extraction followed by a learning based classifier such as SVM, KNN, etc. for performing anomaly detection. The two main drawbacks of using hand-crafted feature-based approaches are the requirement of domain knowledge and poor generalization capability of the features. Deep learning techniques can overcome these challenges while achieving superior performance. In one study, Antipov et al. \cite{Antipov:2015:LVH:2733373.2806332} compared learned features and hand-crafted features for Pedestrian Gender Recognition. The key findings of the research were as follows. The learned features significantly outperformed the hand-crafted features on heterogeneous data composed from different data-sets, with an average mean average precision (mAP) increase of 28.4\% and a maximum increase of as high as 46.5\%. Furthermore, the learned features generalized better than the hand-crafted features on completely unseen data-sets. An average performance improvement of 31.8\% in the mAP value was observed. Finally, they found that smaller CNN models trained from scratch on small data-sets were able to produce compact features that generalized as well as the features produced by much bigger pre-trained networks fine-tuned on the same data-sets.

 Lately, deep learning based approaches for surface anomaly detection have gained a lot of attention. Specifically, the use of Convolutional Neural Networks (CNNs) has seen an exponential increase. One of the primary reasons behind this increased adoption is the ability of CNNs to eliminate the need for domain-specific engineered features by learning complex filters from the training data. In \cite{Masci2012SteelDC}, CNNs were used with max pooling layers for the task of steel defect classification into 7 defect types and their performance was compared with SVM classifiers trained on engineered features. The CNN based approach performed approximately two times better in comparison to the feature descriptor based SVM approach. The best CNN model having 7 hidden layers had an error of 6.79\%, while the best feature i.e. Pyramid of Histograms of Orientation Gradients (PHOG) based classifier had an error of 15.48\%. Another important observation was that the network with large filter sizes did not achieve the best performance. The network with progressively decreasing filter sizes of $ 11\times11, 6\times6$ and $5\times5 $ achieved better performance than the network with $19\times19$ and $13\times13$ filter sizes. Their technique, however, had the following three shortcomings. First, the input image size to the network was restricted to one value because of the use of fully connected layers. Next, the network did only classification without localizing the defect in the images via a pixel-level segmentation mask. Finally, since a large number of kernels per CNN layer were used, this increased the number of network parameters and potentially exposed the models to the problem of over-fitting. Over-fitting is a problem in deep learning where the model is more complicated than is necessary, thereby leading to the memorization of data-sets. Even though an over-fit model can have a good performance on the training data, it does not generalize well and has poor performance on the testing data and does not scale to other testing data.

Another major challenge for the CNN based approaches applied to supervised anomaly detection is that they require a large number of samples of normal and anomalous instances for training \cite{chandola2009anomaly} \cite{chalapathyanomaly}. Specifically, detailed pixel level annotation of the anomalous instances is required for the segmentation tasks. For all practical applications and purposes, this is a major drawback. This is because not only are anomalous instances limited in real-world applications, but also the creation of pixel-level annotated data-sets is cumbersome and expensive. A technique to tackle this challenge was explored on different CNN architectures for surface anomaly detection \cite{WEIMER2016417}. To train the network, the proposed method required sub-sampling of the original images with the extraction of $32\times32$ patches which lead to a 47 fold increment in the number of training samples. However, this sub-sampling approach resulted in extremely long training time of 24 hours and also led to the loss of the global contextual information required for the anomaly detection task. Also, patch based approaches are several times slower than the FCN based approaches where the entire input image is processed in a single forward pass through the network.

A different approach to tackle the challenge of lack of labelled data is to use weakly supervised learning methods. As discussed in the previous section, these techniques try to construct models by learning from weak supervision. One such technique was presented in \cite{compactcnn}, which used a CNN based architecture for the classification and segmentation of anomalies from weakly annotated data. However, their approach had the following shortcomings. The network did not learn to detect the actual shape of the anomaly from the weak labelling. The input size of the image for the model was fixed to $512 \times 512$, which prevented images of other shapes to be fed to the network. It outputted masks of size $128\times128$, which were sixteen times smaller than input image size and resulted in the loss of localization and shape information. This can introduce errors in the calculation of metrics which are based on the shape and size of the defect (anomaly). Also, the network was not tested on any real world data-set which raised concerns regarding its practical application. The paper lacked quantitative analysis for the segmentation task. The qualitative results presented and discussed in the paper showed that the segmentation results were poor, with a lot of false positive pixels i.e., defect free regions classified as defects. The classification part of the model completely relied on the features extracted by the model for the segmentation task. Thus, good segmentation capability was crucial for the classification stage. Lastly, the proposed architecture had approximately 1.13 million parameters. The huge number of parameters made the architecture susceptible to the problem of over-fitting. If the optimization problem is made easier by changing model architectures (by making them deeper and thereby increasing the total number of model parameters), generalization performance can be degraded \cite{kawaguchi2018generalization}.

A network architecture that not only avoids over-fitting but also has a good generalization capability is missing in the literature. In this paper, a novel technique capable of learning to detect the actual shape of the anomalies from not only weakly labeled data but also from a limited number of samples is presented. To achieve this, we explore pre-seeding the preliminary feature extractor with a biologically plausible one.  Empirical testing of the architecture is used to find a compact design.

\section{AnoNet: A fully convolutional network for anomaly detection}
\label{sec:anonet}
AnoNet is a fully convolutional network capable of learning to detect the actual shape of the anomalies not only from weakly labelled data but also from a limited number of instances while achieving state of the art performance.

\subsection{Network Architecture}
Our AnoNet architecture is presented in Figure \ref{fig:anonetoverview}, it is a modification of the CompactCNN \cite{compactcnn}. It is a Fully Convolutional Network (FCN) and therefore overcomes the restriction of a fixed size input faced in the case of CNNs that make use of fully connected layers. AnoNet consists of four convolutional layers and all the layers have a stride of one. For all the layers a zero padding of $\frac{k-1}{2}$ (where $k \times k$ is the kernel size of the layer) is done on all the sides to ensure that the size of the output feature maps are same as the input feature maps. Progressively decreasing filter sizes of $k \times k \;( k \in \{11,7\})$, $7 \times 7$, and $3 \times 3$ are used to allow the network to have a large field of view, which is beneficial for the anomaly detection task. The first layer of AnoNet is the filter layer which is seeded using the Filter Bank initialization technique  (\ref{sec:filterbankinitialization}), which is biologically plausible. The rest of the layers are initialized with a random normal distribution with zero mean and variance of one. The values of the parameters $k$ (filter size) and $n$ (filter stack length) depend on the AnoNet configuration and are summarized in Table \ref{tab:filterexps}. All the layers except the last layer use ReLU activation function which is defined by the Equation \ref{eq:relu}. It can be shown that deep neural networks trained with ReLU train several times faster than their equivalents with tanh units \cite{Krizhevsky:2017:ICD:3098997.3065386}.

\begin{equation}
\label{eq:relu}
    f(x) = max(0,x)
\end{equation}

However, for the segmentation layer (i.e. the last layer),  the tanh activation is used. It was selected since it resulted in a better separation of the anomalies from the normal pixels in comparison to the ReLU and linear activation. Batch Normalization is applied after every layer since it accelerates the training of deep networks by making normalization inherent to the model architecture \cite{batchnormalization}. Normalization of a vector means making it have the mean of zero and a variance of one. For a vector $\textbf{x}$, the normalization equation is given below by Equation \ref{eq:bnequation}, where $E[\textbf{x}]$ is the expectation of $\textbf{x}$ and $Var[\textbf{x}]$ is its variance. 

\begin{equation}
    \label{eq:bnequation}
    \textbf{x} = \frac{(\textbf{x}-E[\textbf{x}])}{\sqrt{Var[\textbf{x}]}}
\end{equation}

\begin{table}[!ht]
\caption{AnoNet: Filter Bank configurations. There are 12 AnoNet configurations in total and their names are given in the configuration column. The filter bank column refers to the filter bank being used.  Filter size column gives the AnoNet parameters $k$ and $n$. The trainable column contains Boolean values. True means that filter layer (first layer) of AnoNet was set to be trainable i.e., its parameters were updated during the training and False means that the parameters were frozen and did not change during the training.}
\begin{center}
\begin{tabular}{cccc}
\toprule
\textbf{Configuration} & \textbf{Filter Bank} & \textbf{Filter Size $(k \times k \times n)$} & \textbf{Trainable} \\ \midrule
LMExp1              & LM                   & 7x7x48                    & False               \\
LMExp2              & LM                   & 7x7x48                    & True              \\
LMExp3              & LM                   & 11x11x48                   & False               \\
LMExp4              & LM                   & 11x11x48                   & True              \\
RFSExp1             & RFS                  & 7x7x38                    & False               \\
RFSExp2             & RFS                  & 7x7x38                    & True              \\
RFSExp3             & RFS                  & 11x11x38                   & False               \\
RFSExp4             & RFS                  & 11x11x38                   & True              \\
SExp1               & S                    & 7x7x13                    & False               \\
SExp2               & S                    & 7x7x13                    & True              \\
SExp3               & S                    & 11x11x13                   & False               \\
SExp4               & S                    & 11x11x13                   & True             \\ \bottomrule
\end{tabular}
\end{center}
\label{tab:filterexps}
\end{table}

In comparison to the segmentation part of CompactCNN \cite{compactcnn} (hereafter referred to as CompactCNN), the AnoNet architecture achieves a massive reduction of 94.29\% in the total number of network parameters from 1.13 million to 64 thousand on average. Despite this huge reduction in parameters, AnoNet outperformed CompactCNN in the anomaly detection task as shown in Section \ref{sec:results}. For a $H \times W$ image, the network outputs a $H \times W$ mask. This is because down-sampling operations such as strided convolutions and pooling have not been performed in the network architecture. This also prevents the artifacts that are introduced during the up-sampling transposed convolution or deconvolution operation. AnoNet has the valuable ability to learn to detect the actual shape of the anomalies from weakly annotated data-sets with a limited number of training samples.

The unique features of the ANoNet architecture are as follow.
\begin{enumerate}
    \item The network is a fully convolutional and does not use strided convolution (i.e., layers with stride $>1$) which does not down-sample the image.  For a $W \times H$ input image, we get a $W\times H$ output mask.  Since the model does not use transposed convolutions for up-sampling, there are no checkerboard artifacts.
    \item The network is shallow and compact which prevents overfitting by design.  Additionally, this allows the training of the network accomplished with only a limited number of training samples.
    \item The compactness of the model causes the size of the intermediate features to be limited which allows the training to be done without having to down-size the image to a lower resolution before making the batches to feed to the GPU.
    \item The model footprint is small which makes it suitable for local execution on edge and IoT devices.
    \item The network can learn to detect the underlying shape of the anomaly despite the weak labelling.
\end{enumerate}

\subsection{Filter Bank Initialization Technique}
\label{sec:filterbankinitialization}
\begin{figure*}[!ht]
\begin{center}
   \includegraphics[width=.8\linewidth]{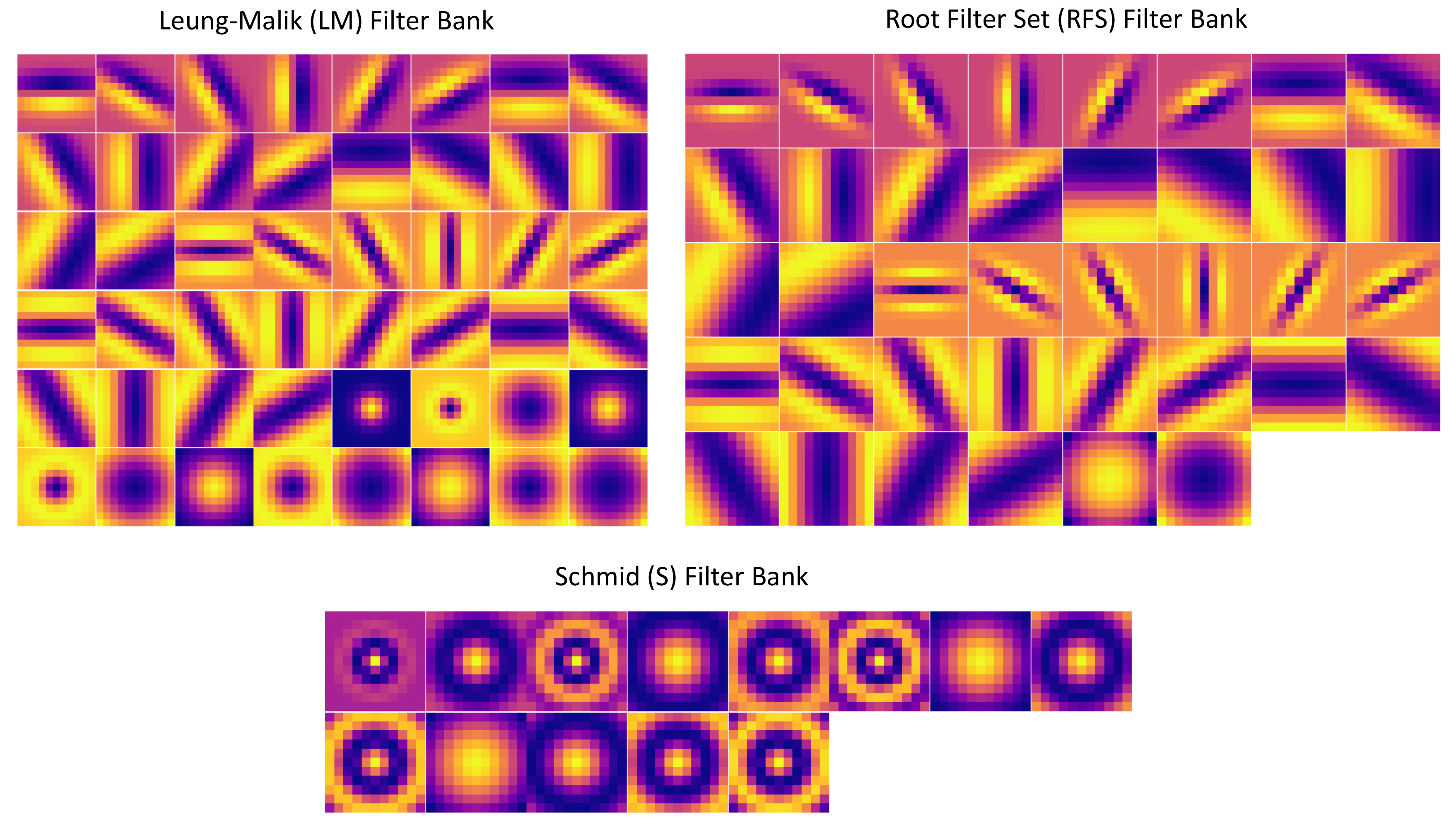}
\end{center}
   \caption{The figure shows the LM, S and RFS filter banks extracted at an 11x11 kernel size. The LM filter bank has a mix of edge, bar and spot filters at multiple scales and orientations. It consists of first and second derivatives of Gaussians at 6 orientations and 3 scales making a total of 36; 8 Laplacian of Gaussian (LOG) filters; and 4 Gaussians. The RFS filter bank consists of 2 anisotropic filters (an edge and a bar filter, at 6 orientations and 3 scales), and 2 rotationally symmetric ones (a Gaussian and a Laplacian of Gaussian). The S filter bank consists of 13 isotropic Gabor like filters \cite{vggfilter}. }
\label{fig:filterbanks}
\end{figure*}

Filter Banks usually refer to a collection of specially designed hand-crafted kernels that are stacked together and applied to images to extract useful features for a particular task. This is usually followed by the use of a learned classifier such as an SVM to perform the classification or segmentation. Gabor filters \cite{gabor}, Wavelet filters \cite{waveletdefect} and Difference of Gaussians are examples of some commonly used filters for texture related tasks. In our proposed technique, three specific filter bank sets namely the Leung-Malik (LM), Schmid (S) and Root Filter Set (RFS) were selected since they contained both rotationally invariant as well as directional filters \cite{Varma03} \cite{Leung2001} \cite{contetbased} \cite{fastgaus}. Thus, they are general and some have claimed that they are biologically plausible \cite{filter}.  Images of each of these three filter banks extracted at an 11x11 kernel size are shown in Figure \ref{fig:filterbanks}.

 For all the experiments, the three filters banks were extracted using the scripts provided in \cite{vggfilter} and are briefly described below.

\begin{enumerate}
    \item \textbf{Leung-Malik (LM) Filter Bank}: The LM filter bank comprises of a set of 48 multi-scale and mutli-orientation filters. There are 36 filters of $1^{st}$ and $2^{nd}$ order derivatives of Gaussians at 6 orientations and 3 scales, along with 8 Laplacian of Gaussian (LOG) filters and 4  Gaussians filters \cite{vggfilter}.
    \item \textbf{Schmid (S) Filter Bank}: The S filter bank comprises of 13 rotationally invariant filters \cite{vggfilter} which have the following form shown in Equation \ref{eq:vgg}.
    \begin{equation}
    \label{eq:vgg}
        F(r,\sigma,\tau) = F_{0}(\sigma,\tau) + cos(\frac{\pi\tau r}{\sigma})\;      e^{-\frac{r^{2}}{2\sigma^{2}}}
    \end{equation}
    The $F_{0}(\sigma,\tau)$ term is added to make the DC component zero. The rotational symmetry of this filter bank can be seen in Figure \ref{fig:filterbanks}.
    \item \textbf{Root Filter Set(RFS) Filter Bank}: As shown in Figure \ref{fig:filterbanks}, the RFS filter bank is similar to the LM filter bank. It comprises of 38 filters and uses a Gaussian and a Laplacian of Gaussian both with $\sigma=10$ pixels, an edge filter at three scales $(\sigma_{x},\sigma_{y}) = \{(1,3),(2,6),(4,12)\}$ and a bar filter at the same three scales \cite{vggfilter}.
\end{enumerate}

\section{Methodology}
Our investigation was conducted in four stages: (1) Analysis of CompactCNN, (2) Visualization Studies, (3) Ablation Studies and (4) AnoNet Filter Bank Studies. 

\subsection{Data-sets}

Four data-sets were selected for experimentation. The data-set selection contained one artificially generated data-set and three real world data-sets all with completely different textures and defects. Each data-set had a limited number of training samples which made the anomaly detection task difficult. Additionally, varying illumination, different camera positions, and orientation added to the complexity. The wide variety and challenging nature of these data-sets ensured that the proposed technique was tested thoroughly and not limited to any particular type of texture and defect. 

\begin{figure*}[!ht]
\begin{center}
   \includegraphics[width=.8\linewidth]{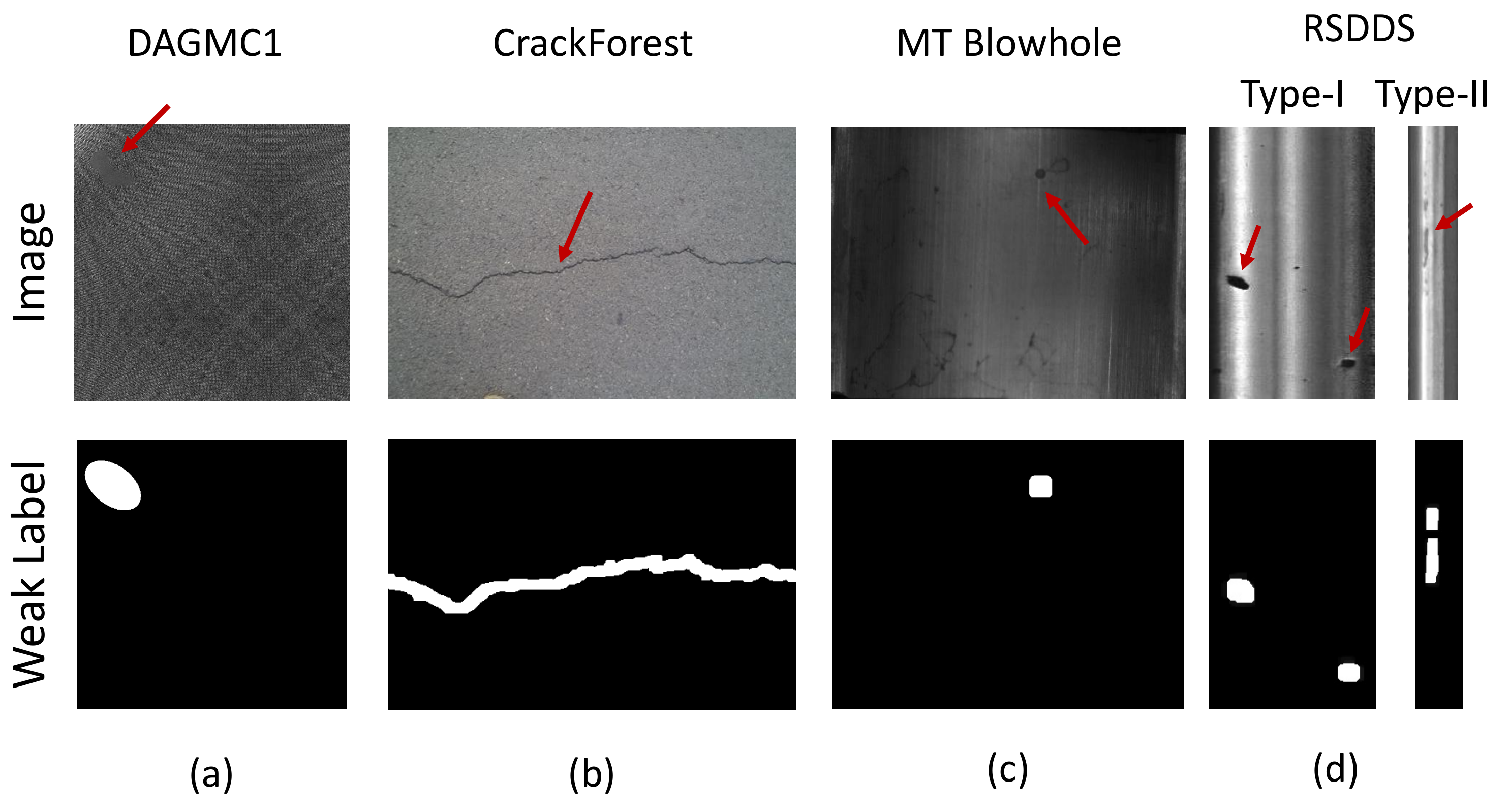}
\end{center}
   \caption{Figure shows one sample image and weakly labelled mask pair per data-set. The red arrows point toward the anomaly in the images. Figure \ref{fig:sampledataset} (a) shows the sample from DAGMC1 having a smudge as the anomaly. Figure \ref{fig:sampledataset} (b) shows a sample from CrackForest data-set and has cracks as the anomaly. Figure \ref{fig:sampledataset} (c) shows the MT Blowhole sample. The defect or anomaly is a Blowhole, a type of surface defect in magnetic tiles. Figure \ref{fig:sampledataset} (d) shows the RSDDS-I and RSDDS-II data-set samples. These have surface defects on express rails and common/heavy haul rails respectively as the anomaly. Dilation was performed for all the data-sets except DAGMC1 by using an $11\times11$ filter to make the masks weakly annotated. The resultant weakly annotated mask examples are shown in this figure.}
\label{fig:sampledataset}
\end{figure*}

The data-sets we used include the following.
\label{sec:datasets}
\begin{enumerate}
    \item \label{ds:dagm} \textbf{DAGM}\cite{dagm} is a synthetic data-set for weakly supervised learning for industrial optical inspection. It contains ten classes of artificially generated textures with anomalies. For this study, the Class\;1 having the smudge defect was selected, since it had the maximum intra-class variance of the background texture of all classes in the data-set. It (hereafter referred to as DAGMC1) contains $150$ images with one defect per image and $1000$ defect free images. For every image a weakly labelled annotation in the form of an ellipse that covers the entire defect is available. The ellipse covers a significant amount of normal texture in addition to the defect making the data-set an excellent test case for loosely labelled data.
    \item \label{ds:crackforest}\textbf{CrackForest} \cite{crackforest} data-set consists of  urban road surface images with cracks as defects. The images contain confounding regions such as shadows, oil spills, and water stains. The images were taken using an ordinary iPhone5 camera. The data-set contains $118$ images and has corresponding pixel level masks for the cracks, all having a size of $320\times480$. The additional confounders along with the limited number of samples available for training make CrackForest another good data-set for anomaly detection evaluation.  
    \item \label{ds:magenetitile}\textbf{Magnetic Tile Defects} \cite{Huang2018} data-set contains images of magnetic tiles collected under varying lighting conditions. Magnetic tiles are used in engines for providing constant magnetic potential. There are five different defect types available, namely Blowhole, Crack, Fray, Break and Uneven. Among these, blowholes and cracks impact the quality of magnetic tiles the most. We use the Blowhole category (referred to as MT\_Blowhole) of this data-set since CrackForest already covers a crack type defect. The Blowhole defect category contains $115$ images of varying sizes and pixel level annotations are available for the defects.
    \item \textbf{RSDDs (Rail surface discrete defects)} \cite{raildefect} is a challenging data-set containing varying sized images of two different types of rails. Rail surface defects are one of the most common and most important forms of failure \cite{raildefect}. Every image contains at least one defect and has a complex background with noise. The RSDDs Type-I category contains 67 images from express rails and the Type-II category contains 128 images captured from common/heavy haul rails. Pixel level annotations are available for the defects for both categories. The heavily skewed aspect ratio of the images and a limited number of training samples make this data-set challenging for anomaly detection.
\end{enumerate}
To ensure that the data-sets had weakly labelled annotations, all the data-sets except DAGMC1 (since it was already weakly labelled) were modified by performing the dilation operation using an $11\times11$ filter. A sample image and weakly annotated mask pair from each data-set are shown in Figure \ref{fig:sampledataset}. 

\subsection{First Stage: Analysis of CompactCNN}
CompactCNN \cite{compactcnn}, a CNN based architecture was presented for the segmentation and classification of anomalies in textured surfaces from weakly annotated data. As discussed in Section \ref{sec:relatedwork}, it failed to learn the actual shape of the anomaly from the weak annotation and could not learn from a limited number of training samples. The following modifications were performed to the segmentation part of this network to overcome its limitation of fixed size input and to investigate a different activation function for the segmentation layer.
\begin{enumerate}
    \item To overcome the restriction of a fixed size input to the network, the fully convolutional segmentation part of the network presented in \cite{compactcnn} was selected and its input was kept as $H\times W \times 1$. (In TensorFlow, None x None x 1 was used in the placeholder to infer the dimension of the tensor from the data.) 
    \item The tanh activation was chosen for the segmentation layer instead of the linear activation used in \cite{compactcnn}, because a linear activation gave poorer segmentation output masks in comparison with the tanh activation. The tanh activation restricted the output to $[-1,1]$, thereby enabling the network to better separate the anomalies from the normal texture. 
\end{enumerate}

\begin{figure*}[!ht]
\centering
\includegraphics[scale=.5]{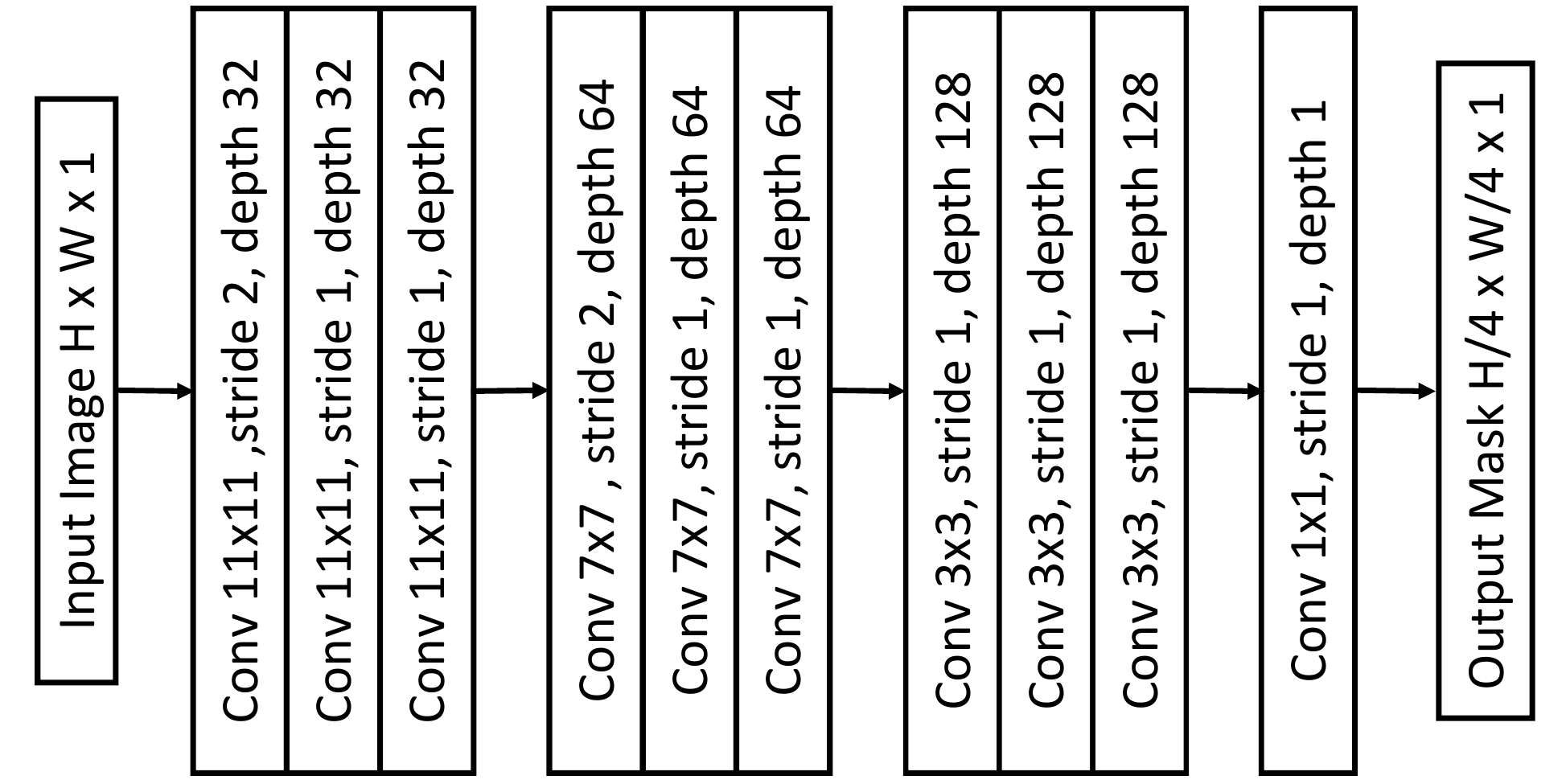}
\caption{The base architecture which is a modified version of the segmentation part of architecture presented by Ra\^cki et al. \cite{compactcnn}. It was pruned through extensive ablation studies to get the AnoNet architecture.}
\label{fig:basenetwork}
\end{figure*}

Initial investigative experiments were conducted using the modified CompactCNN architecture (Figure \ref{fig:basenetwork}) on the DAGMC1 and CrackForest data-set and the results are discussed in Section \ref{sec:results}.

\subsection{Second Stage: Visualization Studies}
\label{visualizaionstudies}
Although deep learning models tend to give superior performance, interpreting why they work is not self-explanatory. To build trustworthy systems and enable their meaningful integration into the industry, model interpretability is important. The deeper the model, the more difficult it is to interpret the results. To check whether the model learnt meaningful filters and analyse why the model failed to learn the underlying shape of the anomalies, the following visualization and activation maximization studies were conducted. 

\renewcommand{\labelenumii}{\roman{enumii}.}
\begin{enumerate}
    \item \textbf{Visualization of intermediate layer outputs:} Observing the intermediate layer outputs gives an understanding of the features being extracted by the kernels for a given input. This study was conducted using the following steps.
    \begin{enumerate}
        \item Forward pass an input image $X$ through the network.
        \item Extract the intermediate activation $a_{i}^{j}(X)$ for $i^{th}$ filter of layer $j$, for all the filters of every layer of the model.
        \item Stack all the intermediate activation outputs and analyse them in a grid.
    \end{enumerate}
    
    \item \textbf{Activation Maximization Study:} Deep learning and human brain analogies are very popular. Certain stimuli can cause specific cells in the brain to have a high response and are known as their preferred stimuli. These preferred stimuli are used in neuro-science to understand the brain. A similar activity for a deep learning model can give us a better understanding of what a neuron or kernel is doing. This technique for deep learning models is known as activation maximization \cite{visualizationsurvey}. Finding an image $X$ that maximizes the activation $a_{k}^{l}(X)$ for $k^{th}$ filter of layer $l$, can be formulated as the following optimization problem.
    \begin{equation}
        X^{*} = \underset{X}{\arg \max}(a_{k}^{l}(X)))
    \end{equation}
    
    The following steps were used for conducting the activation maximization experiments. 
    \begin{enumerate}
        \item Randomly initialize an input image $X$.
        \item Define the optimization loss as the mean value of the activation of a particular filter of a specific layer.
        \item Calculate the gradients of the input image with respect to the loss.
        \item Perform gradient ascent for $n$ steps using a learning rate $\alpha$ on the input image to maximise the activation of the filter. 
    \end{enumerate}
\end{enumerate}

After this stage, the ablation studies were conducted.

\subsection{Third Stage: Ablation Studies}
\label{sec:ablationstudies}
\begin{figure*}[!ht]
\begin{center}
   \includegraphics[width=0.75\linewidth]{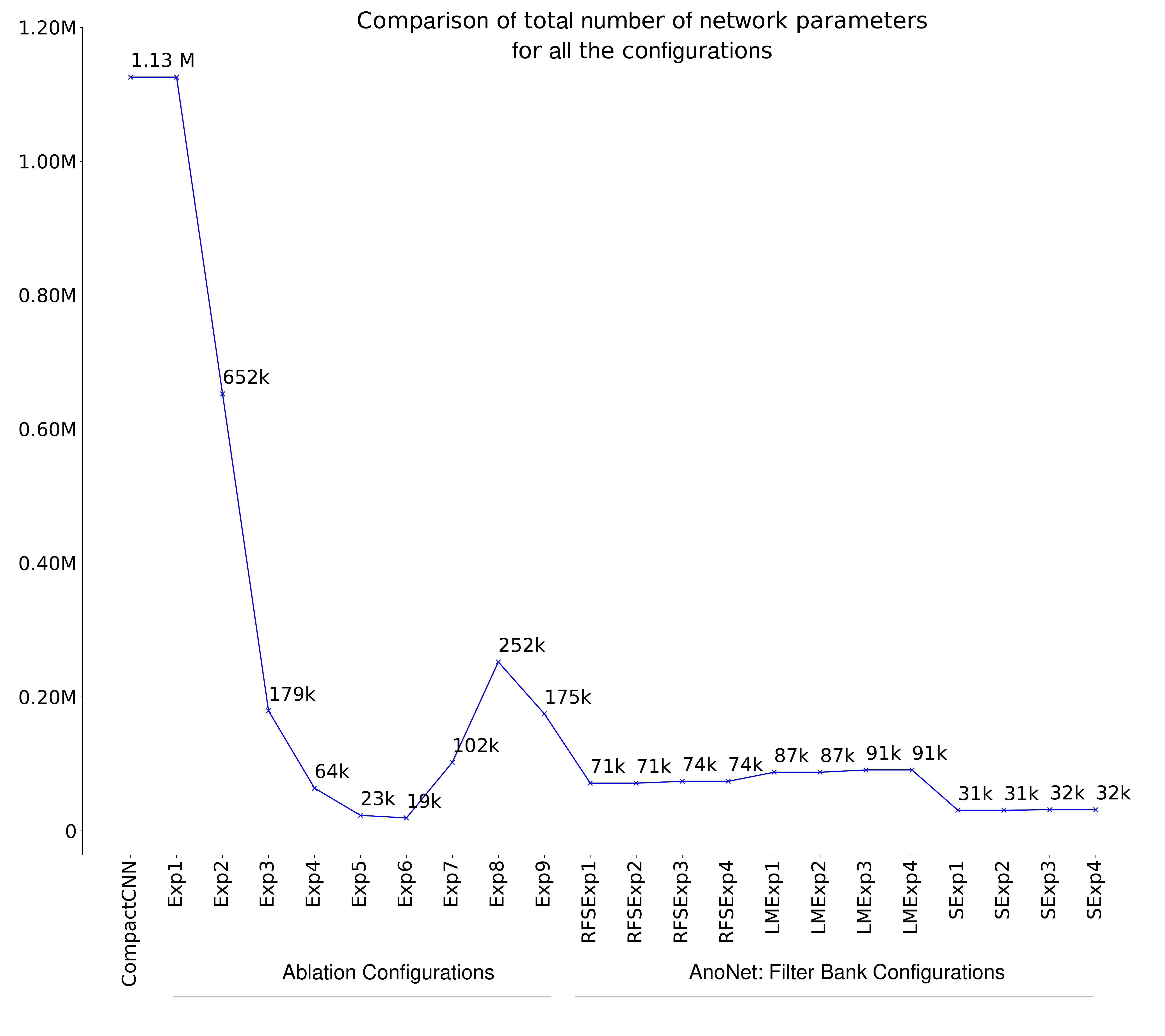}
\end{center}
   \caption{Total network parameter comparison for all the configurations used in the ablation and filter bank studies. The AnoNet architecture achieved a reduction of approximately 94\% in the total number of parameters in comparison to the CompactCNN \cite{compactcnn}. }
\label{fig:exp_parameters}
\end{figure*}

Ablation study is a concept also borrowed from neuro-science and it refers to the selective removal or destruction of tissue to understand its function. In the context of neural networks, ablation studies in the literature are conducted by removing parts, tweaking layers and changing the structure or implementation of neural networks to assess the corresponding changes in the network performance \cite{Lillian2018AblationOA}. Such ablation studies were conducted starting with the modified CompactCNN architecture. In total nine configurations were used for the experiments which are summarized in Table \ref{tab:ablationexps}. Every ablation configuration had three convolutional blocks similar to the network shown in Figure \ref{fig:basenetwork}. In Table \ref{tab:ablationexps}, stride refers to the stride value in the first layer of the first and second blocks of the model respectively. Layers per block imply the number of convolutional layers in the block. Filter sizes are given per block from the first to the third block. Filter depth also follows the same order. For a $H \times W$ input image, for the first two experiments, the output size is $\frac{H}{4}\times\frac{W}{4}$, while for the rest of the experiments, the output size is $H\times W$. To achieve a selective reduction in the number of network parameters, the number of layers per block were gradually reduced from Exp1 to Exp5. We hypothesized that the reduction in the network parameters would address the problem of over-fitting. The distribution of the total number of network parameters for all the configurations is shown in Figure \ref{fig:exp_parameters}. Starting with the ablation configuration Exp2, there was an intended exponential decrease in the number of network parameters. The maximum reduction in parameters was obtained in the Exp6 configuration, with a decrease of 98.3\% in comparison to CompactCNN. Different kernel size combinations were tested in the configurations from Exp6 to Exp9. From the experimental results (Section \ref{sec:results}), Exp4 was found as the optimum configuration and it was selected as the AnoNet architecture. After this selection, we proceeded with the AnoNet Filter Bank studies which are described in the next subsection.

\begin{table*}[!ht]
\caption{Ablation study configurations: There were 9 configurations that had three convolutional blocks each. The configuration column gives the name of the configuration. Stride refers to the stride value in the first layer of the first and second blocks of the model respectively. Layers per block imply the number of convolutional layers in the block. Filter sizes are given per block from the first block to the third block. Filter depth also follows the same order. For the first two experiments, the output size is $\frac{N}{4}\times\frac{N}{4}$ and for the rest of the experiments, the output size is $N\times N$.  }
\begin{center}
\begin{tabular}{ccccc}
\toprule
\textbf{Configuration} & \textbf{Stride} & \textbf{Layers per Block} & \textbf{Filter Sizes} & \textbf{Filter Depth Per Block} \\ \midrule

Exp1                & 2               & 3                         & 11,7,3                                & 32,64,128                                                                   \\
Exp2                & 2               & 2                         & 11,7,3                                & 32,64,128                                                                   \\
Exp3                & 1               & 1                         & 11,7,3                                & 32,64,128                                                                   \\
Exp4                & 1               & 1                         & 11,7,3                                & 32,32,32                                                                    \\
Exp5                & 1               & 1                         & 11,7,3                                & 8,32,32                                                                     \\
Exp6                & 1               & 1                         & 3,3,3                                 & 32,32,32                                                                    \\
Exp7                & 1               & 1                         & 7,7,7                                 & 32,32,32                                                                    \\
Exp8                & 1               & 1                         & 11,11,11                              & 32,32,32                                                                    \\
Exp9                & 1               & 1                         & 3,7,11                                & 32,32,32                                                                    \\ \bottomrule
\end{tabular}
\end{center}
\label{tab:ablationexps}
\end{table*}

\subsection{Fourth Stage: AnoNet Filter Bank Studies}
\label{sec:anonetfilterstudies}
The twelve configurations used for the filter bank studies are presented in Table \ref{tab:filterexps}. The AnoNet architecture used for these studies is discussed in Section \ref{sec:anonet} and shown in Figure \ref{fig:anonetoverview}. The parameter $n$ in this network was set as per the filter bank stack length while the parameter $k$ depended on the filter bank kernel size. The filter banks were extracted at $11\times11$ and $7\times7$ kernel sizes. The extracted filter bank values were used to initialize the weights of the first layer of AnoNet. In Table \ref{tab:filterexps}, the configuration column contains the names of the AnoNet configurations, the filter bank column contains the filter bank type, filter size gives the parameters $k$ and $n$ of AnoNet. The trainable column contains Boolean values where True indicates that the filter layer (first layer) of AnoNet was set to be trainable, i.e., its parameters were updated during the training and False indicates that the parameters were frozen and did not change during the training and thus acted similar to fixed feature extractors. To compare the performance of AnoNet with the state of the art segmentation networks, DeepLabv3 \cite{deeplabv3} was selected as a representative network pre-trained for the semantic segmentation task. The segmentation head of the network was modified according to the anomaly detection segmentation task (to output a binary mask per image) and fine-tuned for all of the data-sets for comparative analysis.

\subsection{Experimental setup}

For the experiments, an NVIDIA Titan Xp graphics card was used. The experiments were conducted using Tensorflow version 1.12. Adadelta optimizer \cite{adadelta} was used with the default settings. The input size to all the network configurations was kept as $(None, None, 1)$. A batch size of 16 was used for all the experiments. All the network weights were initialized as proposed in \cite{He}. All the experiments were conducted for 25 epochs. For the calculation of the F1 score, a threshold value of zero was used across all the experiments. The loss function used for the ablation and AnoNet filter bank experiments was the L2 norm or MSE (Mean Squared Error) which is defined in Equation \ref{eq:mse}.

\begin{equation}
\label{eq:mse}
\displaystyle \operatorname {MSE} ={\frac {1}{n}}\sum _{i=1}^{n}(Y_{i}-{\hat {Y_{i}}})^{2}    
\end{equation}

As discussed in subsection \ref{sec:datasets}, the dilation operation was performed on all the data-sets except DAGMC1, using an $11\times 11$ filter to make the masks weakly annotated. This ensured that normal pixels were included in the masks. The ablation and filter bank experiments were conducted as per the configurations discussed in subsection \ref{sec:ablationstudies} and \ref{sec:anonetfilterstudies} respectively.

\subsection{Evaluation Metrics}
To evaluate the quantitative performance of the models, two metrics were selected. The first metric was the area under curve (AUC) measurement of the receiver operating characteristics (ROC) \cite{aucling}. AUC or AUROC is a reliable measure of the degree or measure of the separability of any binary classifier (binary segmentation masks in this case). It provides an aggregate measure of the model's performance across all possible classification thresholds. An excellent model has AUROC value near to the one and it means that the classifier is virtually agnostic to the choice of a particular threshold. The second metric used for the assessment was the F1 score. It is defined as the harmonic mean of precision (P) and recall (R) and is given by the Equation \ref{eq:f1score}. F1 score reaches its best value at one and the worst score at zero. It is a robust choice for classification tasks since it takes both the false positives and false negatives into account.
\begin{equation}
\label{eq:f1score}
F1 = 2 \times \frac{P \times R}{P + R}    
\end{equation}

\section{Results}
\label{sec:results}
 This section is organised as the discussion of the results of the first to the fourth stages, namely Analysis of CompactCNN, Visualization Studies, Ablation Studies, and AnoNet Filter Bank Studies. It is important to note that for calculating the F1 score, a threshold of zero was used for all the experiments because that is the mean value of the tanh activation's range. The F1 score value can vary depending on the choice of the threshold. However, the AUROC metric takes into account all the possible thresholds into its calculation. Since the DAGMC1 data-set contains defect free examples, AUROC values cannot be calculated for this data-set.

\subsection{First Stage: Analysis of CompactCNN}
\begin{figure}[!ht]
\begin{center}
   \includegraphics[width=.85\columnwidth]{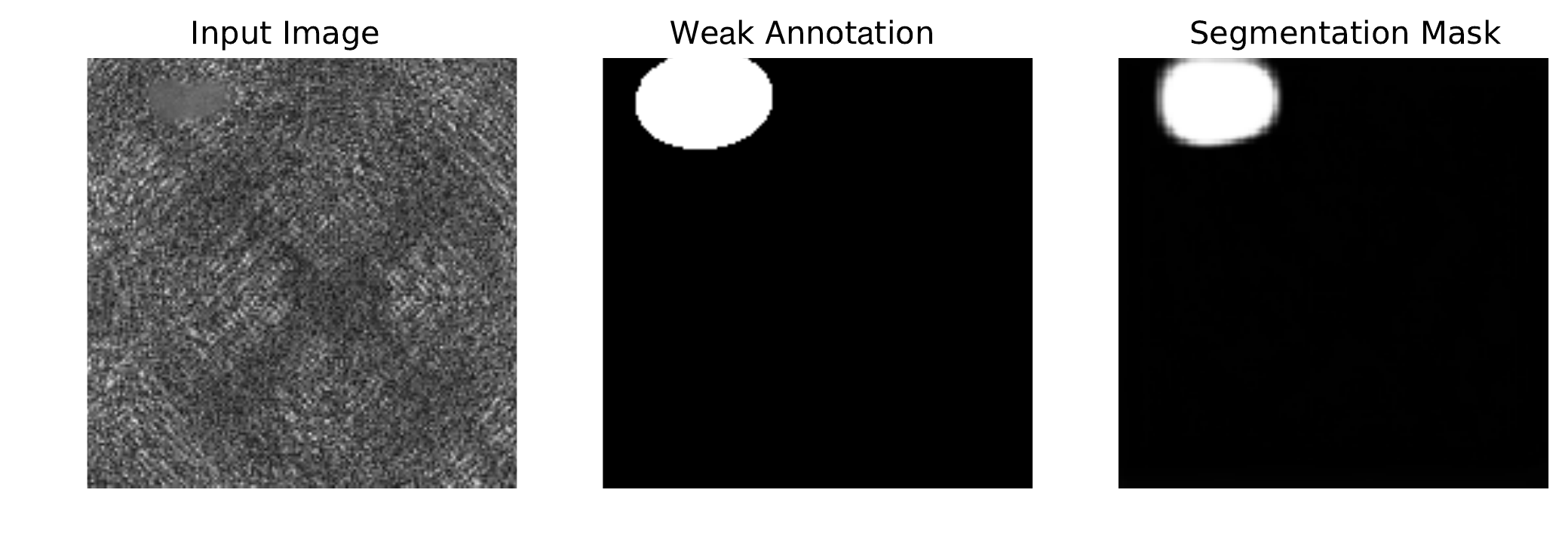}
\end{center}
   \caption{ The segmentation mask shows the output of the modified CompactCNN architecture over-fitting to the DAGMC1 data-set. It fails to learn the underlying shape of the anomaly from the weak annotation. For the input image, it outputs a segmentation mask similar to the ellipse shaped annotation used for training the network.}
\label{fig:overfitting}
\end{figure}

Experiments were conducted on the DAGMC1 data-set using the modified CompactCNN architecture. The results showed that the modifications led to a significant improvement in the F1 score from 0.04 to 0.97 (a threshold of zero was used). It produced better qualitative segmentation results than the ones presented in \cite{compactcnn}. However, even for the modified architecture, the segmentation shape of the anomalous region was oval and over dilated just like the weakly labelled masks used for training and is shown in Figure \ref{fig:overfitting}. To test and check whether the architecture works on real-world data-sets, experiments were conducted on the CrackForest data-set \cite{crackforest}. The model achieved an impressive F1 score of $0.901$. Next, to further analyse why the model failed to learn the underlying shape of the anomalies and get a deeper understanding of the learnt features, visualization studies were conducted and the results are discussed next.

\subsection{Second Stage: Visualization Studies}
\begin{figure}[!ht]
\begin{center}
   \includegraphics[width=0.85\columnwidth]{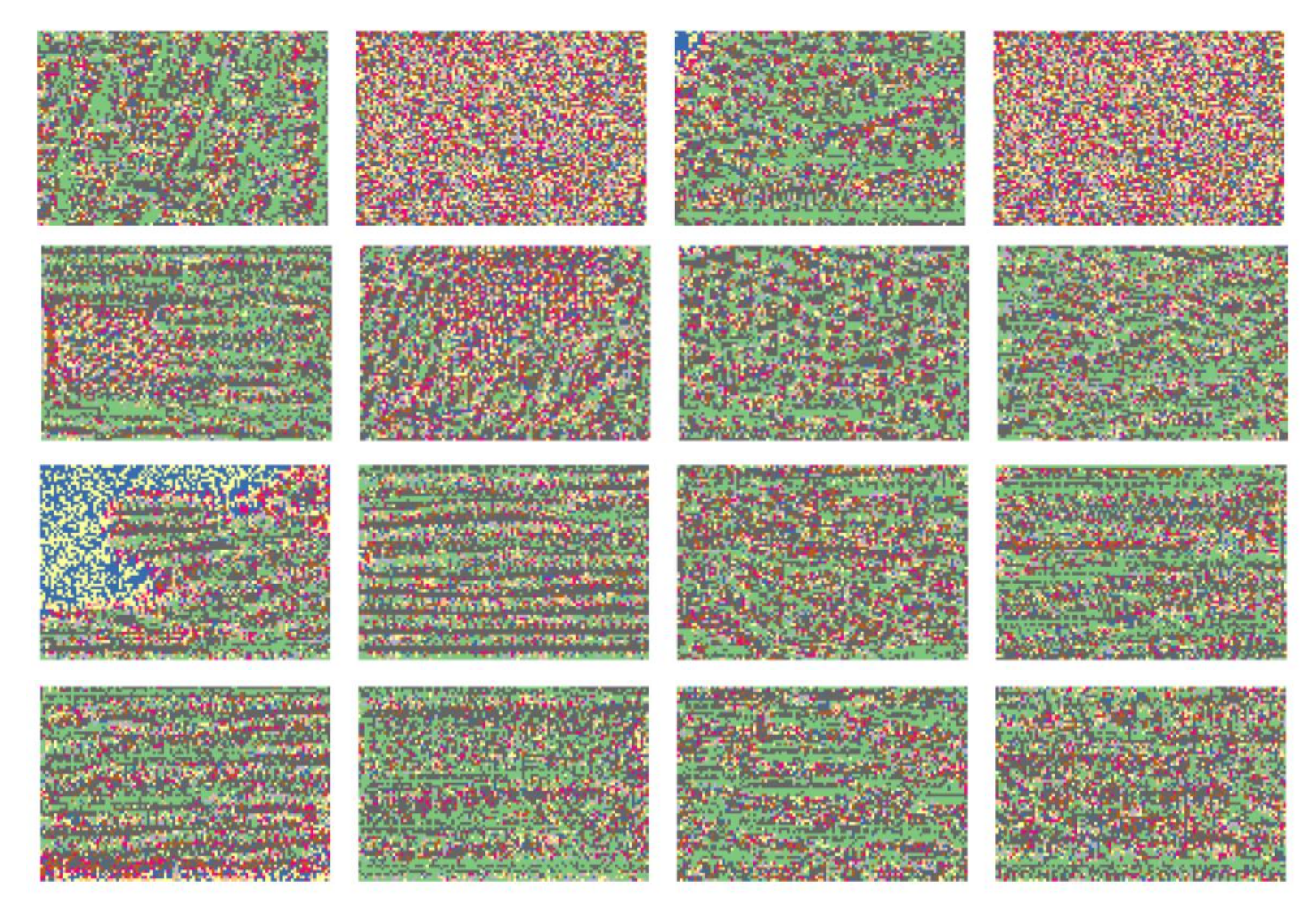}
\end{center}
   \caption{Few randomly selected activation maximization results for the modified CompactCNN trained on the CrackForest data-set. The key observations from these images were that the resultant texture looked like cracks (the anomalies in the CrackForest data-set) and all of the patterns looked similar. The green pixels indicate high intensity values and other colors indicate low intensity values. Each image shows one of the possible input patterns obtained after the gradient ascent optimization which maximised the output for a specific filter of a particular layer. Since the patterns looked like cracks, this showed that the network was looking for cracks in the images for performing the segmentation. Most of the resultant images looked similar but with some random rotations. This pointed towards that the network had redundancy and was possibly over-fitting to the data-set. The noisy examples without any crack like texture are the ones that did not converge from the random initialization. (Best viewed in colour.)}
\label{fig:activationmax}
\end{figure}

 The results of the intermediate layer visualization and activation maximization studies are discussed below.
\begin{enumerate}
    \item \textbf{Visualization of intermediate layer outputs:} A few random samples of the intermediate layer activation study conducted for the model trained on the CrackForest data-set are shown in Figure \ref{fig:interoutput}. From these feature visualizations, we found that the initial layers were not extracting anything useful since most of them were black. The second observation was that most of the filters were looking for similar features in the later layers. This pointed towards the possibility that the model had a lot of redundancy and was over-fitting.

\begin{figure}[!ht]
\begin{center}
\includegraphics[width=0.85\columnwidth]{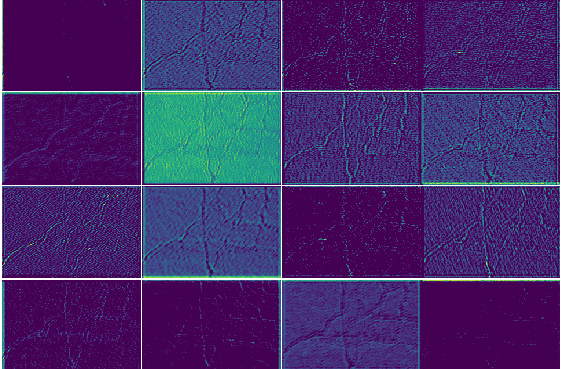}
\end{center}
 \caption{Few randomly selected samples of intermediate feature visualization for the modified CompactCNN trained on the CrackForest data-set. The key observation from these images is that most of the intermediate features extracted by the network looked similar. This pointed towards potential redundancy in the network since most of the learnt filters were looking for similar patterns. (Best viewed in colour.)}
\label{fig:interoutput}
\end{figure}
    \item \textbf{Activation Maximization Study:} In this study, the activation maximization was done for 500 steps for every filter of the modified CompactCNN model trained on the CrackForest data-set. It is important to note here that the activation maximization results are not unique. The process was conducted ten times for every kernel with step sizes of 50 and 100 and the results lead to the same observation. The preferred stimuli for the kernels looked like cracks which showed that these were looking for the right kind of inputs. A few examples of visualizations obtained using the activation maximization approach are shown in Figure \ref{fig:activationmax}. An analysis of the preferred stimuli for all the filters in the model showed that most of the filters were looking for similar patterns which were rotated by some random angle. This observation caused us to conclude that the network was possibly over-fitting and there was redundancy in the network. Because of this observation, we also hypothesized that making the initial filters rotation invariant could be potentially helpful for the anomaly detection task. This led to the inclusion of the filter bank to replace the preliminary network layers.  
\end{enumerate}

After these visualization studies, to validate that the network was over-fitting, we proceeded with the ablation studies, the results of which are discussed below.
 
 \subsection{Third Stage: Ablation Studies}
\begin{figure*}[!ht]
\begin{center}
   \includegraphics[width=\linewidth]{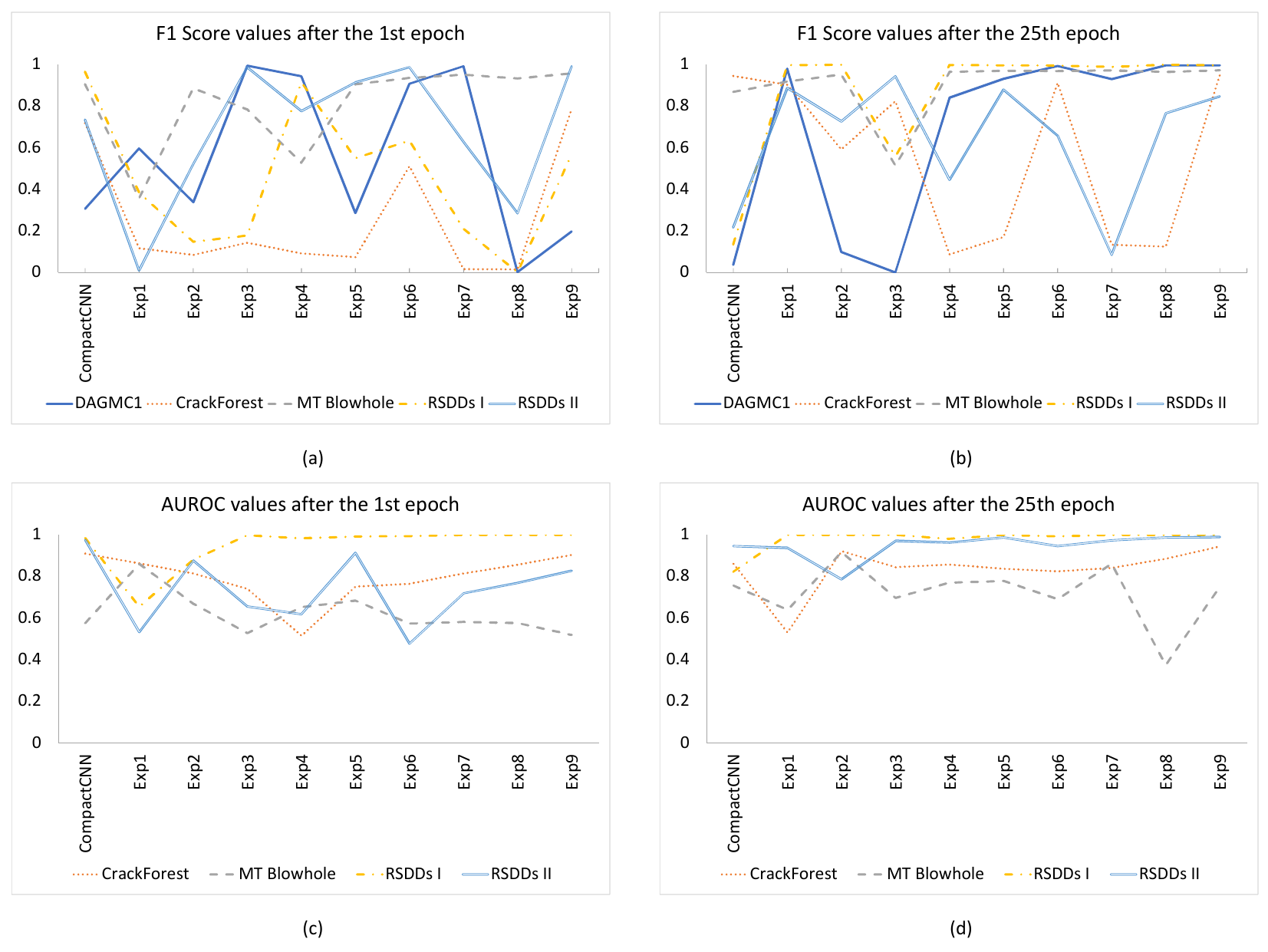}
\end{center}
   \caption{F1 score and AUROC values for all the configurations of the ablation experiments. Figures \ref{fig:ablationgraphs} (a) and (b) show the F1 score values for all the configurations after the first epoch and twenty-fifth epoch respectively. Figures \ref{fig:ablationgraphs} (c) and (d) show the AUROC values for all the configurations after the first epoch and twenty-fifth epoch respectively. As can be seen from the graphs, after the first epoch the metric values were lower in comparison to the values after the twenty-fifth epoch. (Best viewed in colour.)}
\label{fig:ablationgraphs}
\end{figure*}

\begin{figure*}[!ht]
\begin{center}
   \includegraphics[width=\linewidth]{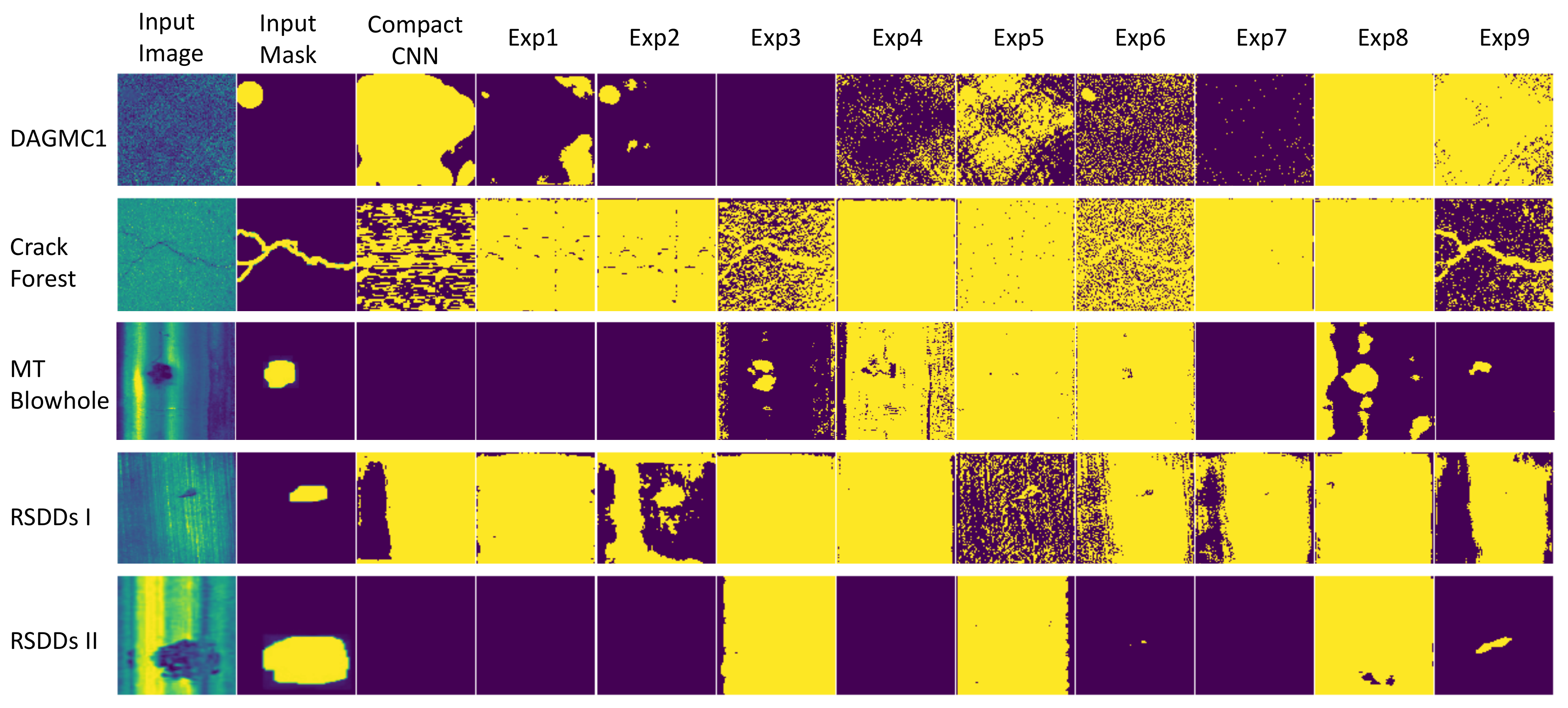}
\end{center}
   \caption{Sample segmentation outputs for the ablation experiments after the first epoch. Almost all of the configurations failed to output meaningful segmentation masks after the first epoch. (Best viewed in colour.)}
\label{fig:ablationimagesm1}
\end{figure*}

\begin{figure*}[!ht]
\begin{center}
   \includegraphics[width=\linewidth]{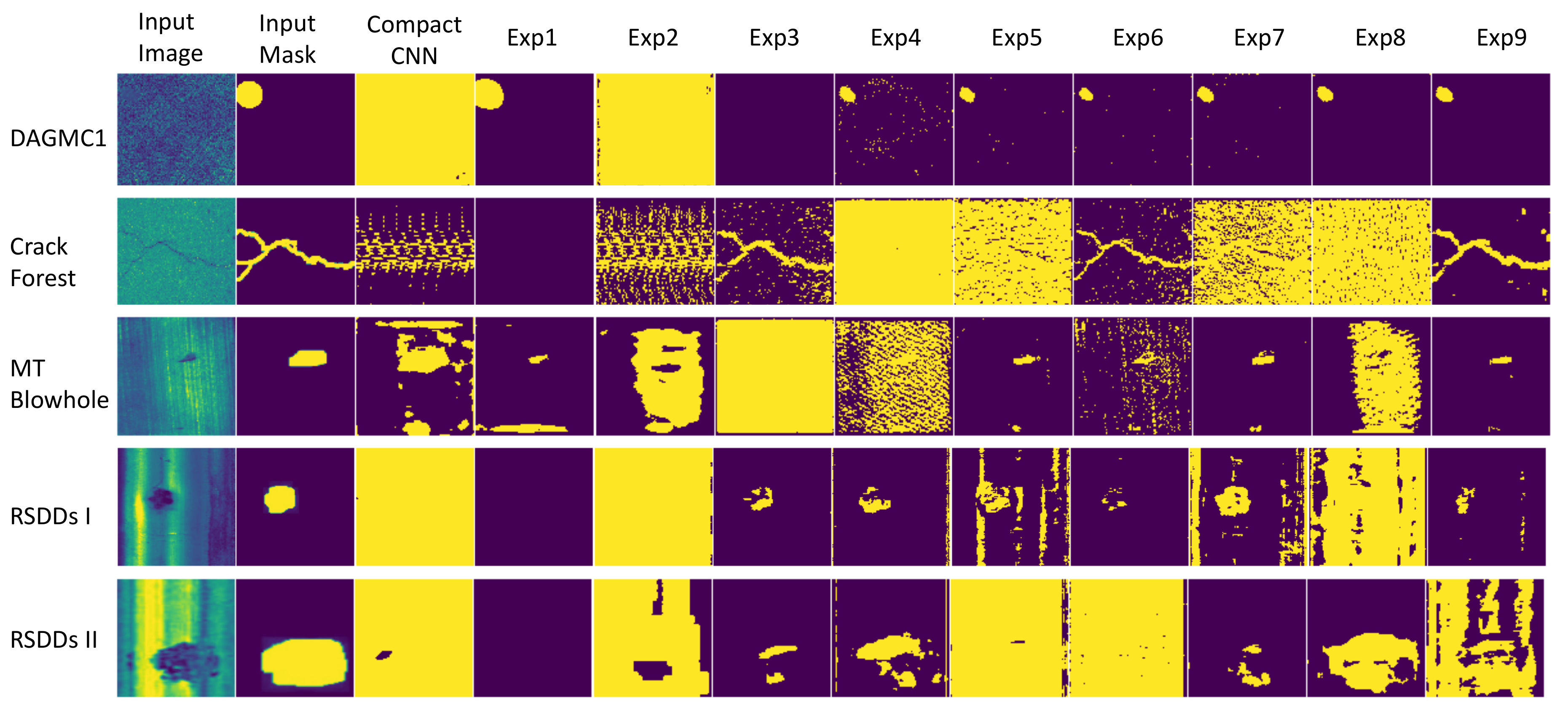}
\end{center}
   \caption{Sample segmentation outputs for the ablation experiments after the twenty-fifth epoch. The configurations learnt to output meaningful segmentation masks but the segmentation quality was poor. Interestingly, the configurations from Exp4 onward learnt to output the actual shape of the anomaly from the weakly labelled training data. (Best viewed in colour.) }
\label{fig:ablationimagesm25}
\end{figure*}

The results of the ablation experiments for all the nine configurations along with the CompactCNN architecture are shown in Figure \ref{fig:ablationgraphs}. There are 4 graphs in total which capture the F1 score and AUROC values for all the data-sets for every configuration. The F1 score graphs are shown in Figure \ref{fig:ablationgraphs} (a) and \ref{fig:ablationgraphs} (b) and  while the AUROC values are in Figures \ref{fig:ablationgraphs} (c) and \ref{fig:ablationgraphs} (d) respectively. For every data-set, one random sample from the validation set was chosen to show sample segmentation outputs. Figures \ref{fig:ablationimagesm1} and \ref{fig:ablationimagesm25} show the segmentation output of the networks after the first and twenty-fifth epoch respectively. As can be seen in Figure \ref{fig:ablationimagesm1}, for all the configurations across all the data-sets, the models failed to output meaningful segmentation masks after the first epoch. This is in concurrence with the lower F1 score and AUROC values of the graphs in Figure \ref{fig:ablationgraphs} (a) and (c) respectively. After the $25^{th}$ epoch, the models learnt to output meaningful segmentation masks that localized the anomaly. This can be seen by the higher metric values in Figure \ref{fig:ablationgraphs} (b) and (d) as well as from the sample segmentation outputs shown in Figure \ref{fig:ablationimagesm25}. Starting with Exp4, all the configurations learnt to identify the actual shape of the anomaly for the DAGMC1 data-set, which can be seen in the first row of Figure \ref{fig:ablationimagesm25}. This confirmed our hypothesis that the modified CompactCNN network was overparameterized which caused the problem of over-fitting. Exp4 configuration was found to have the best trade-off between performance and the total number of network parameters. It achieved a striking 94.3\% reduction from 1.13 million network parameters to only 64 thousand, in comparison to the CompactCNN. Additionally, on an average across all the five data-sets, it achieved a performance improvement of 51.62\% to an F1 score of 0.67 and a 5.44\% improvement to an AUROC value of 0.89. The Exp4 configuration was therefore selected as the AnoNet architecture. Subsequently, the AnoNet Filter Bank studies were conducted and its results are discussed in the next subsection. 

\subsection{Fourth Stage: AnoNet Filter Bank Studies}
\begin{figure*}[!ht]
\begin{center}
   \includegraphics[width=\linewidth]{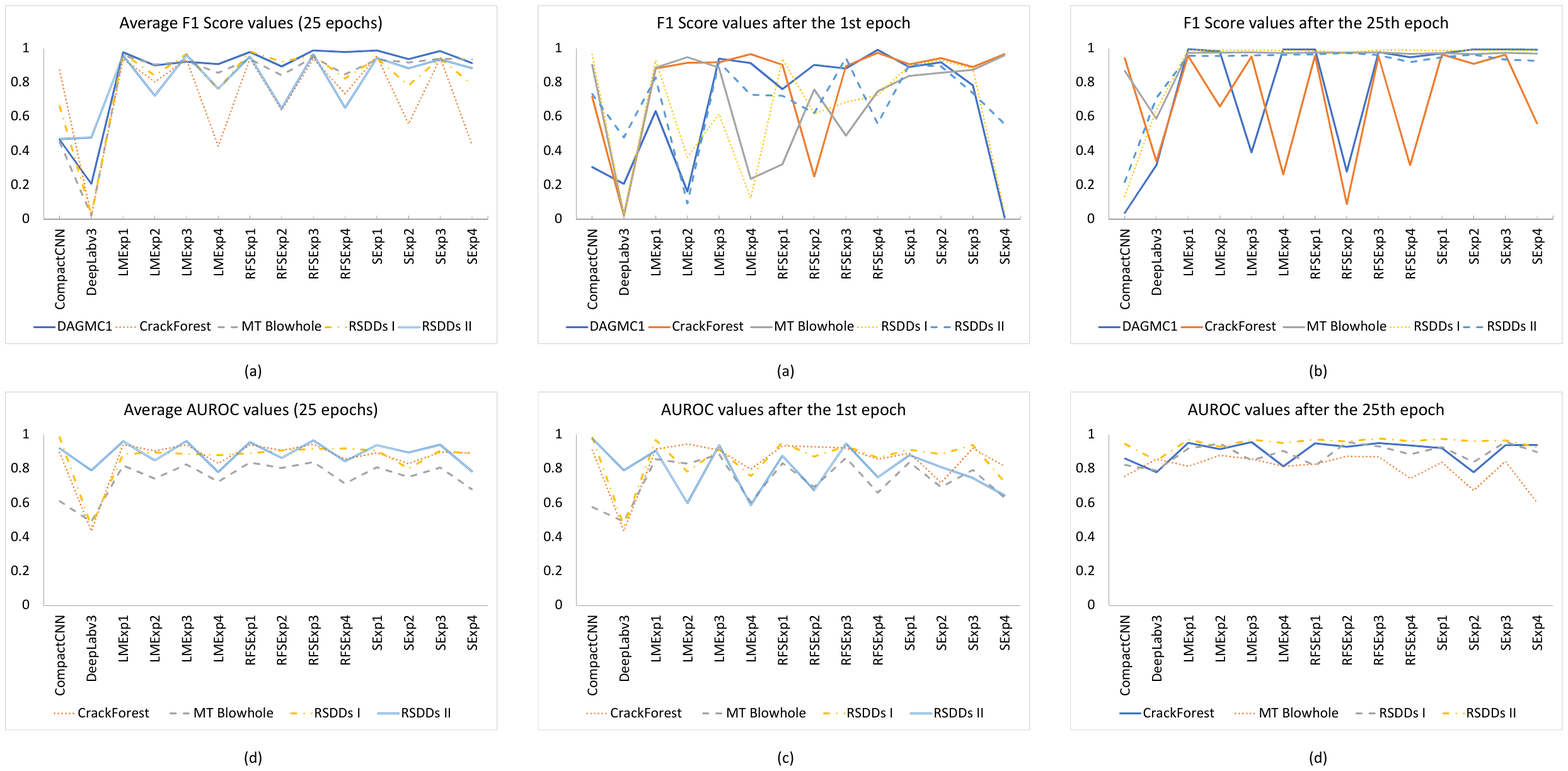}
\end{center}
   \caption{F1 score and AUROC values for all the configurations of the AnoNet filter bank experiments compared to the CompactCNN and DeepLabv3. Figures \ref{fig:ablationgraphs} (a) and (b) show the F1 score values for all the configurations after the first epoch and twenty-fifth epoch respectively. Figures \ref{fig:ablationgraphs} (c) and (d) show the AUROC values for all the configurations after the first epoch and twenty-fifth epoch respectively. All the odd numbered filter bank configurations seemed to perform better than the even number configurations. The SExp1 configuration on an average performed the best across all the data-sets. (Best viewed in colour.)}
\label{fig:filtergraphs}
\end{figure*}

\begin{figure*}[!ht]
\begin{center}
   \includegraphics[width=\linewidth]{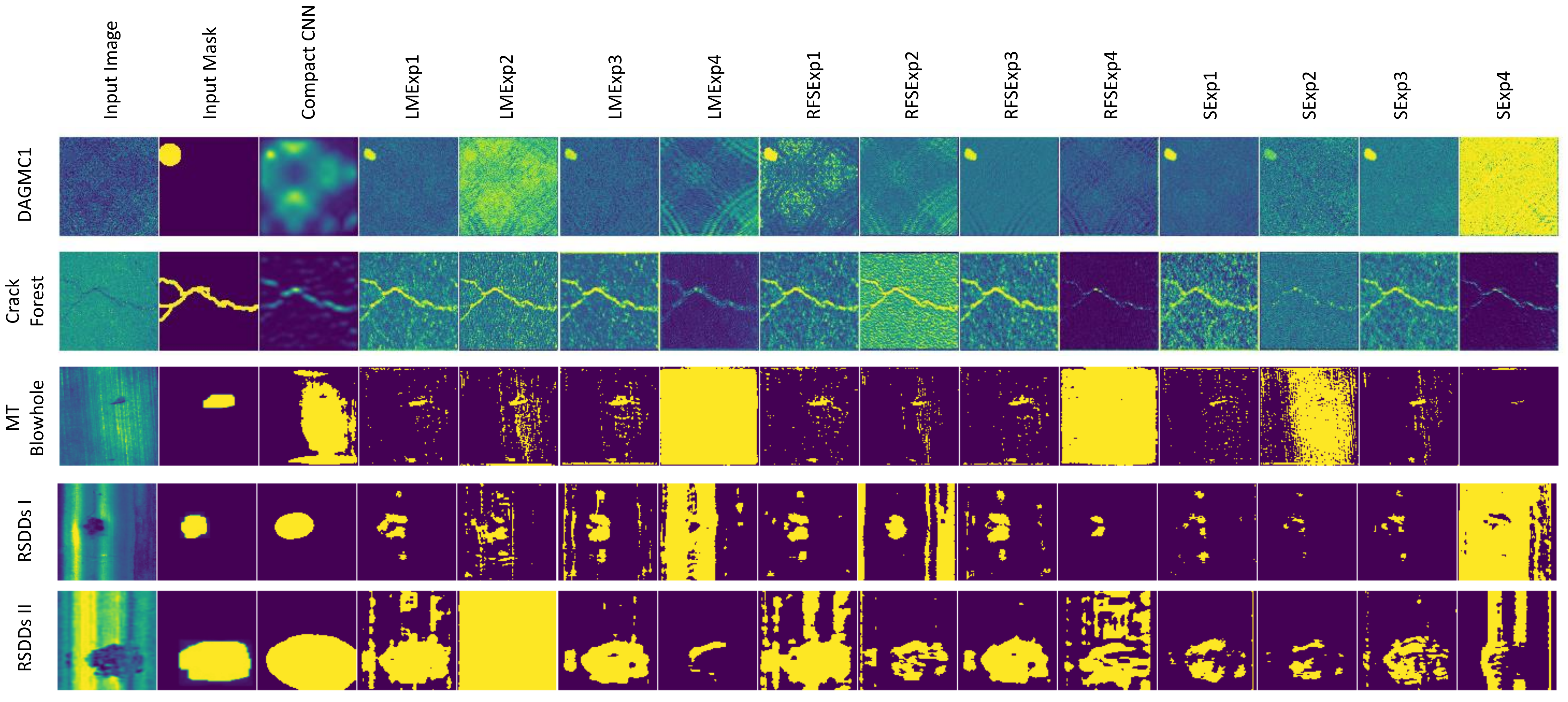}
\end{center}
   \caption{Sample segmentation outputs for filter experiments after the first epoch. As can be seen from the outputs, all the odd numbered configurations learnt to detect anomalies after the first epoch. They also learnt to detect the actual shape of the anomalies from the weak labels. (Best viewed in colour.)}
\label{fig:filterimagesm1}
\end{figure*}

\begin{figure*}[!ht]
\begin{center}
   \includegraphics[width=\linewidth]{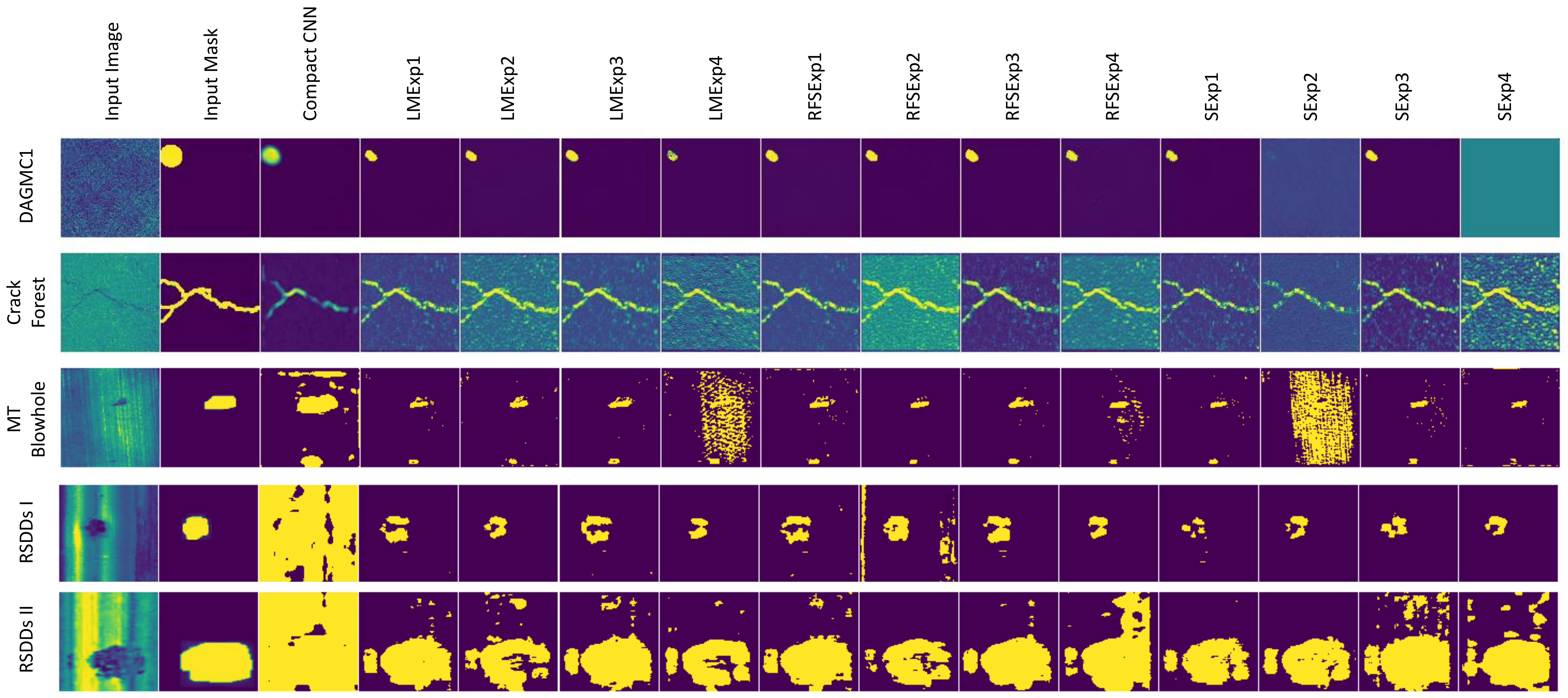}
\end{center}
   \caption{Sample segmentation outputs for filter experiments after 25 epochs. Almost all the AnoNet configurations were outputting the actual shape of the anomalies. The separation of the anomaly and normal pixels had increased for AnoNet. CompactCNN over-fit to the weak annotations and did not learn to detect the actual shape of the anomaly. (Best viewed in colour.)}
\label{fig:filterimagesm25}
\end{figure*}

 The results of the twelve AnoNet Filter Bank configurations in comparison to the CompactCNN and the DeepLabv3 architectures are presented in Figure \ref{fig:filtergraphs}. Similar to the ablation results, the F1 score graphs are shown in Figure \ref{fig:filtergraphs} (a) and \ref{fig:filtergraphs} (b) and  while the AUROC values are in Figures \ref{fig:filtergraphs} (c) and \ref{fig:filtergraphs} (d) respectively. The segmentation outputs after the $1^{st}$ and $25^{th}$ epoch are shown in Figures \ref{fig:filterimagesm1} and \ref{fig:filterimagesm25} respectively. Interestingly, as it can be seen from Figure \ref{fig:filterimagesm1}, all the odd numbered configurations across filters learnt to output the actual shape of the anomaly just after the first epoch for all the five data-sets. The same thing can be observed from the high F1 score and AUROC value of these configurations in the graphs of Figure \ref{fig:filtergraphs} (a) and (c) respectively. In concurrence with our expectations, the rotationally invariant S filter bank performed better than the directional LM and RFS filter banks. This was even though the S filter bank had only 13 filters in comparison to the 48 and 38 filters of LM and RFS filter banks respectively. It is possible that the rotational invariance of the S filter bank allowed it to extract good features across varying data-sets with different texture and defect types leading to its overall best performance.

To measure the overall average performance of the models across all the data-sets, we used AvgF1AUROC which is calculated as follows.

\begin{enumerate}
    \item Calculate the average of F1 scores across all the data-sets for every configuration.
    \item Calculate the average of AUROC values across all the data-sets for every configuration.
    \item Find the average of values calculated in Step 1 and Step 2, to find the AvgF1AUROC value for every configuration.
\end{enumerate} 

The AvgF1AUROC metric gave the average performance of the network for all the data-sets by equally weighing the F1 score and AUROC values. Remarkably, after only a single epoch the SExp1 configuration achieved the highest value AvgF1AUROC of 0.884, followed by LMExp3 with a value of 0.8835. The configurations that had the filter layers frozen during training performed better than the ones that allowed parameter update for the filter layers. After 25 epochs, RFSExp3 had the best performance with an AvgF1AUROC value of 0.952, even though it had a poorer performance after the first epoch. The best configuration for every data-set based on F1 Score, AUROC and average of F1 Score and AUROC value after the $1^{st}$ epoch and the $25^{th}$ epoch are given in Table \ref{tab:filterexpsbestconfig}. We see that some of the configurations which achieved the best performance for individual data-sets had their weights set to trainable. It is interesting to see that even though the SExp1 configuration performed the best on an average for all data-sets, it was not the best performer individually. In comparison to CompactCNN, the SExp1 configuration achieved a performance improvement of 11.5\% to an AvgF1AUROC value of 0.884 after the $1^{st}$ epoch and a 46.4\% improvement to an AvgF1AUROC value of 0.942 after the $25^{th}$ epoch. While the improvement in the AvgF1AUROC score in comparison to DeepLabv3 after the $1^{st}$ epoch was of 153.9\% and 40.8\% after the $25^{th}$ epoch respectively.  Additionally, AnoNet also had a massive 92.2\% reduction in the total number of parameters from 1.13 million to 64 thousand with respect to the CompactCNN. The DeepLabv3 had around 60 million parameters which is approximately 937 times more than AnoNet. The detailed performance comparison of AnoNet with CompactCNN after the $1^{st}$ and $25^{th}$ epoch for all the data-sets is given in Tables \ref{tab:anonetvscompactcnn1epoch} and \ref{tab:anonetvscompactcnn25epoch} respectively. AnoNet performed better than CompactCNN and DeepLabv3 across all the data-sets. All these performance improvements were although AnoNet outputted 16 times more pixel values per image in comparison to the CompactCNN. We also compared AnoNet performance with state-of-the-art techniques for Road Crack Detection. AnoNet outperforms all the methods on the CrackForest data-set which can be seen from Table \ref{tab:roadcrackcomparison}.

\begin{table}[!ht]
\caption{Best AnoNet configurations for every data-set based on F1 Score, AUROC value and average of F1 Score and AUROC value after the 1st and the 25th epoch.}
\begin{center}
\begin{tabular}{lcccccc}
\toprule
\multirow{2}[5]{1.5cm}{data-set}            & \multicolumn{3}{c}{After 1st epoch} & \multicolumn{3}{c}{After 25th epoch} \\ \cmidrule(lr){2-4} \cmidrule(lr){5-7}
             & F1 Score  & AUROC   & Average & F1 Score  & AUROC    & Average \\ \midrule
DAGMC1      & RFSExp4   & N.A.    & N.A.    & LMExp1    & N.A.     & N.A.    \\
CrackForest & RFSExp4   & LMExp2  & LMExp2  & SExp1     & LMExp3   & LMExp3  \\
MT Blowhole & SExp4     & LMExp3  & LMExp2  & LMExp3    & LMExp2   & LMExp2  \\
RSDDs I     & RFSExp1   & LMExp1  & LMExp1  & RFSExp3   & SExp3    & SExp3   \\
RSDDs II    & RFSExp3   & RFSExp3 & RFSExp3 & RFSExp2   & RFSExp3  & RFSExp3 \\ \bottomrule
\end{tabular}
\end{center}
\label{tab:filterexpsbestconfig}
\end{table}

\begin{table}[!ht]
\caption{Comparison of AnoNet, CompactCNN and DeepLabv3 after the 1st Epoch.}
\begin{center}
\begin{tabular}{lcccccc}

\toprule
\multirow{2}{*}{data-set} & \multicolumn{2}{c}{AnoNet (Proposed Method)} & \multicolumn{2}{c}{CompactCNN} & \multicolumn{2}{c}{DeepLabv3} \\ \cmidrule(lr){2-3} \cmidrule(lr){4-5} \cmidrule(lr){6-7}
                        & \multicolumn{1}{c}{F1 Score}     & \multicolumn{1}{c}{AUROC}       & \multicolumn{1}{c}{F1 Score}     & \multicolumn{1}{c}{AUROC} & \multicolumn{1}{c}{F1 Score}     & \multicolumn{1}{c}{AUROC}         \\ \midrule
DAGMC1      & \textbf{0.991} & N.A.           & 0.305 & N.A.  & 0.207 & N.A.  \\
CrackForest & \textbf{0.973} & \textbf{0.945} & 0.717 & 0.910 & 0.020 & 0.436 \\
MT Blowhole & \textbf{0.958} & \textbf{0.883} & 0.904 & 0.576 & 0.022 & 0.491 \\
RSDDs I     & \textbf{0.964} & \textbf{0.984} & 0.939 & 0.969 & 0.027 & 0.465 \\
RSDDs II    & \textbf{0.944} & \textbf{0.976} & 0.734 & 0.947 & 0.478 & 0.791 \\ \bottomrule
\end{tabular}
\end{center}
\label{tab:anonetvscompactcnn1epoch}
\end{table}

\begin{table}[!ht]
\caption{Comparison of AnoNet, CompactCNN and DeepLabv3 after the 25th Epoch.}
\begin{center}
\begin{tabular}{lcccccc}
\toprule
\multirow{2}{*}{data-set} & \multicolumn{2}{c}{AnoNet (Proposed Method)} & \multicolumn{2}{c}{CompactCNN} & \multicolumn{2}{c}{DeepLabv3} \\ \cmidrule(lr){2-3} \cmidrule(lr){4-5} \cmidrule(lr){6-7}
                        & \multicolumn{1}{c}{F1 Score}     & \multicolumn{1}{c}{AUROC}       & \multicolumn{1}{c}{F1 Score}     & \multicolumn{1}{c}{AUROC} & \multicolumn{1}{c}{F1 Score}     & \multicolumn{1}{c}{AUROC}         \\ \midrule
DAGMC1      & \textbf{0.995} & N.A.           & 0.036 & N.A.  & 0.315 & N.A.  \\
CrackForest & \textbf{0.964} & \textbf{0.956} & 0.944 & 0.861 & 0.338 & 0.780 \\
MT Blowhole & \textbf{0.977} & \textbf{0.878} & 0.867 & 0.756 & 0.588 & 0.856 \\
RSDDs I     & \textbf{0.990} & \textbf{0.958} & 0.134 & 0.823 & 0.641 & 0.791 \\
RSDDs II    & \textbf{0.972} & \textbf{0.977} & 0.216 & 0.946 & 0.709 & 0.848  \\ \bottomrule

\end{tabular}
\end{center}
\label{tab:anonetvscompactcnn25epoch}
\end{table}

\begin{table}[!ht]
\caption{Comparison of AnoNet with the road crack detection systems on the CrackForest data-set.}
\begin{center}
\begin{tabular}{ll}
\toprule
Method                          & F1 Score \\ \midrule
AnoNet (Proposed Method)        & \textbf{0.9734}     \\ 
Canny                           & 0.3073                 \\
CrackTree \cite{cracktree}                  & 0.7089                 \\
CrackIT \cite{crackit}                & 0.7164                 \\
CrackForest (KNN) \cite{crackforest}              & 0.7944                 \\
CrackForest (SVM) \cite{crackforest}              & 0.8571                 \\
CrackForest (One-Class SVM) \cite{crackforest}     & 0.8377                 \\
Structred Prediction using CNNs \cite{structuredprediction} & 0.9244                 \\ \bottomrule  
\end{tabular}
\end{center}
\label{tab:roadcrackcomparison}
\end{table}

Finally, to analyse how the choice of loss function impacts the network performance, experiments were conducted. Preliminary results from experiments conducted using the ablation experiment configurations on the CrackForest data-set using CrossEntropy (H) as defined by Equation \ref{eq:bce} and mean squared error (MSE) (defined by the Equation \ref{eq:mse}) as the two loss functions show that the MSE loss worked better than CrossEntropy. In comparison to the models trained using the CrossEntropy loss, the models trained using MSE loss, on an average, achieved a 44.5\% higher F1 score value of 0.71 and 6.88\% higher AUROC value of 0.85.

\begin{equation}
    \label{eq:bce}
    H = -\frac{1}{n}\sum_{i=1}^{n} [y_{i}\log(\hat{y}_{i})+(1-y_{i})log(1-\hat{y}_{i})]
\end{equation}

\section{Conclusion}
We have presented AnoNet, a fully convolutional architecture (Figure \ref{fig:anonetoverview}) for anomaly detection in textured surfaces using weakly labelled data which outputs a $H\times W$ segmentation mask for a $H\times W$ input image. This prevents the loss of localization of the anomaly with respect to the original image. The network's asset is the ability to learn to output the actual shape of the anomaly despite not only the weak annotations but also the limited data. In comparison to the CompactCNN \cite{compactcnn}, on an average, the AnoNet architecture achieves a humongous 94\% reduction in the total number of network parameters from 1.13 million to 64 thousand while achieving better performance. This not only results in the reduction of the computational complexity of the model thereby reducing inference time but also overcomes the challenge of over-fitting by design. A unique filter bank based initialization technique for AnoNet is presented. To the best of our knowledge, no such work has been done for weakly supervised anomaly detection in textured surfaces. Experiments conducted on four challenging data-sets showed that, in comparison to CompactCNN and DeepLabv3, AnoNet achieved an impressive improvement in performance on an average across all data-sets by 106\% to an F1 score of 0.98 and by 13\% to an AUROC value of 0.942. This performance improvement was even though AnoNet predicted 16 times more pixels per image in comparison to CompactCNN. The model learnt to detect anomalies after just one epoch. AnoNet has the advantage that it can learn from a limited number of images. In the experiments conducted for the RSDDs-I data-set, it learnt to detect anomalies after a single pass through only 53 training images. Currently, there is no bench-marking available for weakly supervised anomaly detection. We hope that AnoNet serves as a benchmark for future studies. For future work on AnoNet, investigations need to be conducted on how the choice of filter banks and the trainable parameter for the filter bank layer affects the performance of AnoNet on different types of textures and anomalies. Since the ground truth itself is not accurate in weakly labelled anomaly detection, the Intersection over Union (IoU) metric is another possible way for measuring the quantitative performance. Additionally, it would be interesting to see whether AnoNet achieves similar performance on data-sets with more than one defect type per image.

\bibliographystyle{unsrt}  
\bibliography{references}  

\end{document}